\documentclass{article}

\usepackage{microtype}
\usepackage{graphicx}
\usepackage{subfigure}
\usepackage{booktabs} 

\usepackage{hyperref}
\usepackage{pifont}

\usepackage[accepted]{icml2024}

\usepackage{amsmath}
\usepackage{amssymb}
\usepackage{mathtools}
\usepackage{amsthm}

\usepackage[capitalize,noabbrev]{cleveref}

\theoremstyle{plain}

\theoremstyle{definition}

\theoremstyle{remark}

\usepackage[textsize=tiny]{todonotes}

\usepackage{url}
\usepackage{amsfonts}
\usepackage{nicefrac}
\usepackage{microtype}
\usepackage{xcolor}
\definecolor{darkgreen}{rgb}{0.1,0.7,0.1}
\definecolor{darkblue}{rgb}{0.1,0.1,0.7}
\hypersetup{
colorlinks,
linkcolor=red,
anchorcolor=blue,
urlcolor=darkblue,
citecolor=darkblue,
}

\usepackage{colortbl}
\usepackage{pifont}
\usepackage{bbding}   
\usepackage{fontawesome}  
\usepackage{color}
\usepackage[algo2e,ruled,noend,linesnumbered]{algorithm2e}
\usepackage{soul}
\usepackage{multirow}
\usepackage{makecell}
\newcommand{\cmark}{\ding{51}}
\newcommand{\xmark}{\ding{55}}

\newcommand{\ie}{\textit{i}.\textit{e}.}
\newcommand{\eg}{\textit{e}.\textit{g}.}
\usepackage{caption}
\definecolor{ForestGreen}{RGB}{34,139,34}
\definecolor{brickred}{rgb}{0.8, 0.25, 0.33}
\definecolor{lavender}{RGB}{230, 230, 250}
\definecolor{LemonChiffon}{RGB}{255, 240, 205}
\definecolor{MistyRose}{RGB}{255, 228, 225}
\usepackage{rotating}  

\usepackage{pifont}
\newcommand{\hollowstar}{\text{\ding{72}}}
\newcommand{\tri}{\text{\ding{115}}}
\newcommand{\dia}{\text{\ding{117}}}
\usepackage{bm}
\usepackage[misc]{ifsym}
\usepackage{wrapfig}
\newcommand{\fade}[1]{{\color{gray!60}{}#1{}}}
 
\usepackage{textcomp}  
\usepackage{scalerel}  
\def\fire{\scalerel*{\includegraphics{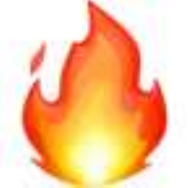}}{\textrm{\textbigcircle}}}
\def\snowflake{\scalerel*{\includegraphics{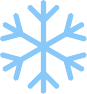}}{\textrm{\textbigcircle}}}

\usepackage{tikz}
\newcommand{\rotbox}[1]{\rotatebox{60}{#1}}

\icmltitlerunning{CrossGET: Cross-Guided Ensemble of Tokens for Accelerating Vision-Language Transformers}

\begin{document}

\twocolumn[
\icmltitle{CrossGET: \underline{Cross}-\underline{G}uided \underline{E}nsemble of \underline{T}okens for \\ Accelerating Vision-Language Transformers}



\icmlsetsymbol{correspondence}{$\dagger$}

\begin{icmlauthorlist}
\icmlauthor{Dachuan Shi}{thu,shlab}
\icmlauthor{Chaofan Tao}{hku}
\icmlauthor{Anyi Rao}{stanford}
\icmlauthor{Zhendong Yang}{thu}
\icmlauthor{Chun Yuan}{thu,correspondence}
\icmlauthor{Jiaqi Wang}{shlab,correspondence}
\end{icmlauthorlist}

\icmlaffiliation{thu}{Tsinghua University}
\icmlaffiliation{shlab}{Shanghai AI Laboratory}
\icmlaffiliation{hku}{The University of Hong Kong}
\icmlaffiliation{stanford}{Stanford University}

\icmlcorrespondingauthor{Chun Yuan}{yuanc@sz.tsinghua.edu.cn}
\icmlcorrespondingauthor{Jiaqi Wang}{wjqdev@gmail.com}

\icmlkeywords{Multimodal Model, Vision-Language Transformers, Model Acceleration, Token Ensemble}

\vskip 0.3in
]



\printAffiliationsAndNotice{}  

\begin{abstract}
  Recent vision-language models have achieved tremendous advances. However, their computational costs are also escalating dramatically, making model acceleration exceedingly critical.
To pursue more efficient vision-language Transformers, this paper introduces \textbf{Cross}-\textbf{G}uided \textbf{E}nsemble of \textbf{T}okens (\textbf{\emph{CrossGET}}), a general acceleration framework for vision-language Transformers. This framework adaptively combines tokens in real-time during inference, significantly reducing computational costs while maintaining high performance. \textit{CrossGET} features two primary innovations:
1) \textit{Cross-Guided Matching and Ensemble}. \textit{CrossGET} leverages cross-modal guided token matching and ensemble to effectively utilize cross-modal information, achieving wider applicability across both modality-independent models, \eg, CLIP, and modality-dependent ones, \eg, BLIP2.
2) \textit{Complete-Graph Soft Matching}. \textit{CrossGET} introduces an algorithm for the token-matching mechanism, ensuring reliable matching results while facilitating parallelizability and high efficiency. Extensive experiments have been conducted on various vision-language tasks, such as image-text retrieval, visual reasoning, image captioning, and visual question answering. The performance on both classic multimodal architectures and emerging multimodal LLMs demonstrates the framework's effectiveness and versatility. The code is available at \href{https://github.com/sdc17/CrossGET}{https://github.com/sdc17/CrossGET}.

\end{abstract}

\section{Introduction}

The AI community is currently witnessing the bloom of vision-language models \cite{kiros2014unifying, karpathy2014deep, antol2015vqa, vinyals2015show, yang2016stacked, huang2017instance, radford2021learning, wang2022unifying, li2022blip, li-etal-2023-lavis}, with Transformer-based models such as CLIP \cite{radford2021learning}, BLIP/BLIP2 \cite{li2022blip, li2023blip}, and GPT-4V \cite{openai2023gpt4vision} emerging as prominent in recent research.
These models are capable of tackling a broad range of vision-language tasks, such as Image-Text Retrieval \cite{https://doi.org/10.48550/arxiv.1509.04942}, Vision Reasoning \cite{suhr2018corpus}, Image Captioning \cite{lin2014microsoft}, and Visual Question Answering \cite{antol2015vqa}. 
Nevertheless, the notable improvement is at the expense of significantly increased computational cost, making it less accessible for consumers with limited resources.

The computational cost of Transformers increases monotonically with the input tokens. Token reduction, which reduces the number of tokens processed during forward, is an effective strategy to mitigate high computational costs for both vision Transformers \cite{rao2021dynamicvit, liang2022not, bolya2022tome} and language Transformers \cite{goyal2020power, wang2021spatten, kim2022learned}. Although studied on the acceleration of unimodal models, a non-negligible research gap persists in multimodal contexts. 

In vision-language transformers, a straightforward idea involves leveraging cross-modal information to guide token reduction. This concept can be intuitively applicable in modality-independent frameworks, such as CLIP~\cite{radford2021learning}. However, recent popular vision-language Transformers like BLIP/BLIP2 \cite{li2022blip, li2023blip} and LLaVA \cite{liu2023visual, liu2023improved} are modality-dependent, with the vision encoder processing first. While vision features can guide token reduction in the following networks, incorporating language priors to guide token reduction in the vision encoder poses a challenge. The issue of facilitating bidirectional guidance, which would allow the preceding modality to benefit from information in the succeeding modality for token reduction, remains an open question.

\begin{figure*}[tb]
    \centering
    \captionsetup{font={small}}
    \includegraphics[width=0.8\linewidth]{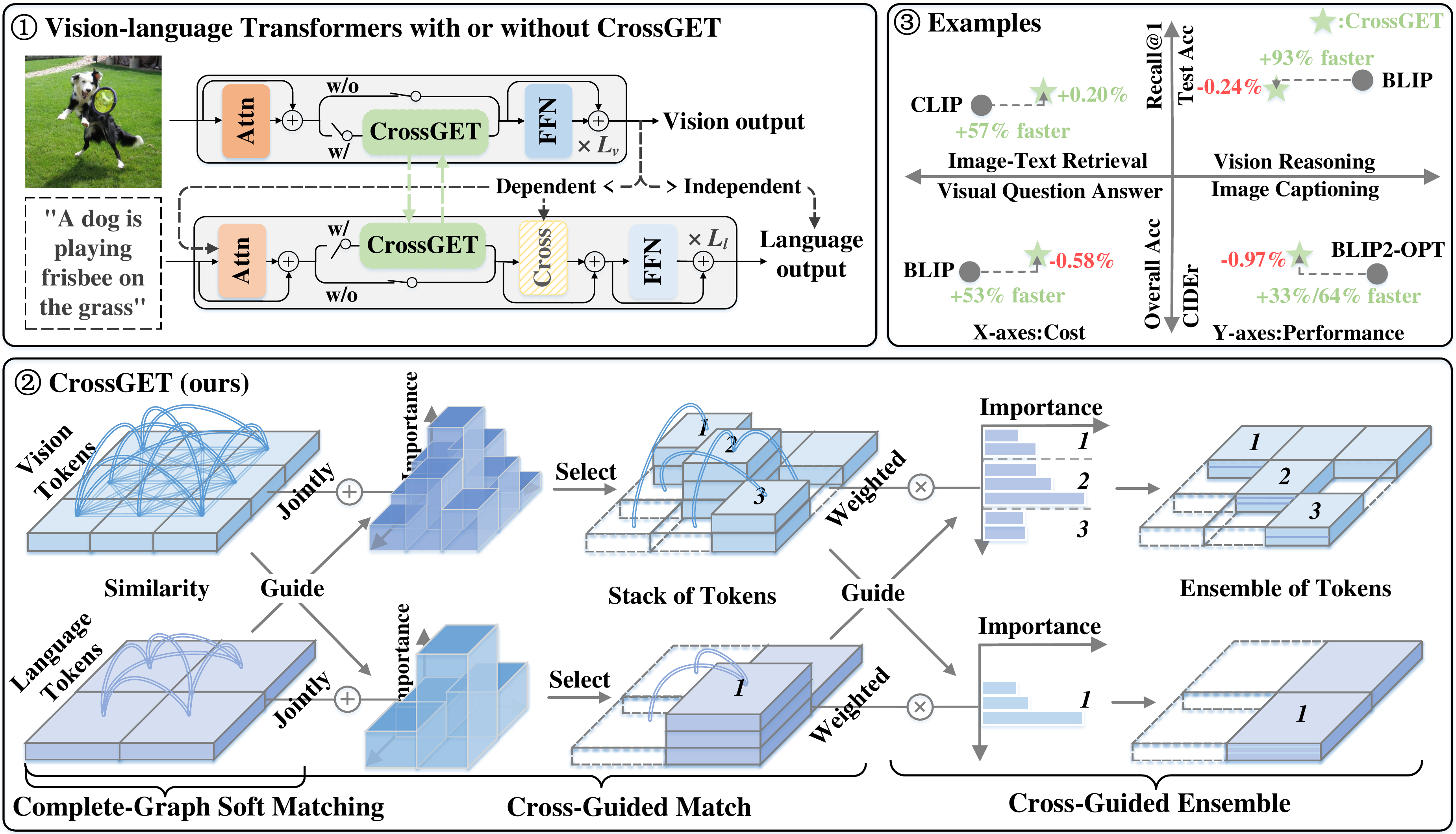}
    \caption{\textbf{Overview of \textit{CrossGET}.} \textbf{\ding{172}} \textit{CrossGET} is a general multimodal token reduction framework that applies to both modality-independent and modality-dependent models. \textbf{\ding{173}} \textit{CrossGET} jointly considers the token similarity derived from intra-modal complete-graph soft matching and the token importance indicated by cross-modal guidance to determine which tokens should be combined. The cross-modal importance is subsequently utilized to weight tokens within each stack and output their ensembles. \textbf{\ding{174}}  Compared with the original models, \textit{CrossGET} achieves considerable computation saving and acceleration with negligible performance degradation.}
    \label{figure overview}
    \vspace{-10pt}
\end{figure*}

This paper introduces \textit{CrossGET}, a general acceleration framework designed to efficiently reduce the number of tokens for both  \textit{modality-independent} and \textit{modality-dependent} vision-language Transformers with bidirectional guidance. \textit{CrossGET} features two primary innovations: \textit{cross-guided matching and ensemble} and \textit{complete-graph soft matching}. 

Firstly, \textit{CrossGET} utilizes \textit{cross-guided matching and ensemble} to identify and ensemble redundant tokens, which applies to both modality-independent and modality-dependent models. \textit{CrossGET} incorporates cross tokens into both vision and language branches to facilitate learning of cross-modal importance and to guide the selection of redundant tokens.\footnote{A naive solution is to calculate the similarity between vision and language tokens directly. However, in modality-dependent models, different modality branches are aligned crosswise or sequentially, rendering cross-modal similarity inaccessible to preceding branches. \textit{CrossGET} enables cross tokens within each modality to act as proxies for other modalities, allowing the preceding modality to leverage information from the succeeding modality without being constrained by the order of calculations.} Secondly, for the underlying mechanism of token matching, \textit{CrossGET} formulates it as a discrete optimization problem and proposes an approximate algorithm \textit{complete-graph soft matching} to secure reliable matching results while maintaining parallelizability for high efficiency. The contributions of this paper are summarized as follows:

\begin{figure*}[tb]
    \centering
    \captionsetup{font={small}}
    \includegraphics[width=1.0\linewidth]{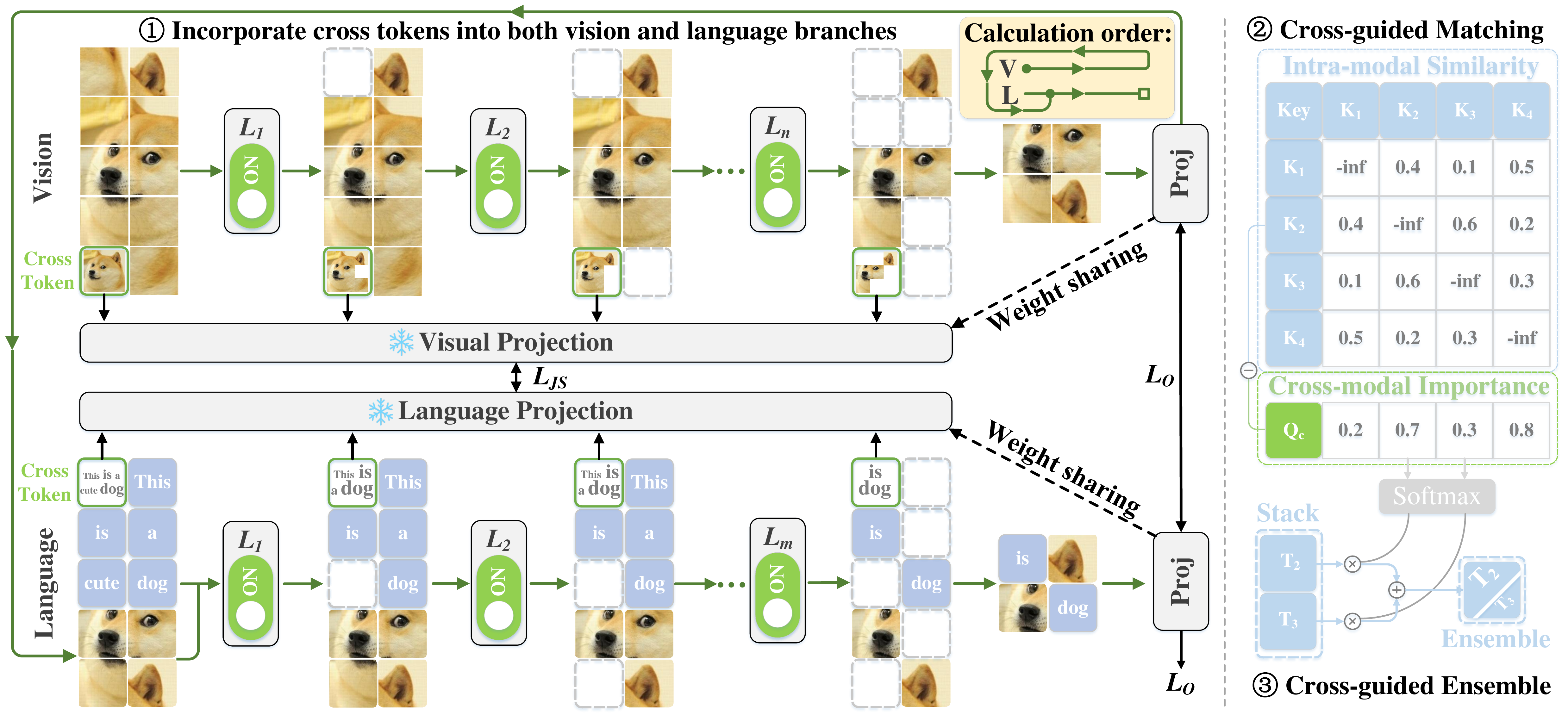}
    \caption{\textbf{Diagram of introducing and leveraging cross-model guidance for vision-language Transformers.} \textbf{\ding{172}} Cross tokens learn cross-modal information by closing the after-projection distance between cross tokens of different modalities. The switches indicate that it is free to choose whether to reduce tokens in different modalities and layers. \textbf{\ding{173}} Cross tokens provide cross-modal importance as a metric to guide token matching. \textbf{\ding{174}} The metric also guides the weighted summation of the stacked tokens to produce token ensemble results.}
    \label{figure cross}
    \vspace{-6pt}
\end{figure*}

\begin{itemize}
\vspace{-8pt}
\item It is one of the pioneering efforts in token ensemble framework for vision-language Transformers, achieving general applicability across both modality-independent and modality-dependent models. The approach is also validated in zero-shot scenarios.

\item It introduces \textit{cross-guided matching and ensemble}, a novel approach for effectively leveraging cross-modal information. It also proposes a \textit{complete-graph soft matching} algorithm for reliable token-matching results while maintaining parallelizability.

\item Its versatility has been validated across a broad range of vision-language tasks, datasets, and model architectures. This is also the \textit{first} application of token ensemble approaches to the modality-dependent pipeline of BLIP2~\cite{li2023blip}, which is a widely adopted paradigm among recent large vision-language Transformers, \eg, LLaVA~\cite{liu2023visual}, MiniGPT-4~\cite{zhu2023minigpt}, and mPLUG-Owl~\cite{ye2023mplug}.
\end{itemize}

\section{Related Work}

\textbf{Vision-Language Transformers} \ According to the dependency on calculation order across different modalities, existing vision-language Transformers can be classified into two main categories: 1) Modality-independent models \cite{li2020oscar, li2021align, radford2021learning, kim2021vilt, singh2022flava}. For example, CLIP \cite{radford2021learning} is a representative model. These models allow for both the visual and language branches to be calculated simultaneously. 2) Modality-dependent models \cite{li2021align, yu2022coca, li2022blip, alayrac2022flamingo}, exemplified by BLIP-based models \cite{li2022blip} and BLIP2/LLaVA-based \cite{li2023blip, zhu2023minigpt, dai2023instructblip, liu2023visual, gao2023llama} multimodal LLMs. In these models, calculation must commence with the visual branch, as the language branch relies on outputs from the visual branch as part of its inputs. \textit{CrossGET} applies to both modality-independent and modality-dependent scenarios. 

\textbf{Model Acceleration Techniques} Numerous model acceleration techniques exist, for example, knowledge distillation \cite{hinton2015distilling, zhang2019your, jiao2019tinybert, wang2020minilm, touvron2021training, yang2022masked}, model pruning \cite{han2015deep, he2017channel, fan2019reducing, zhu2021vision, chavan2022vision, tao2023structured}, and quantization \cite{xiao2022smoothquant, tao2022compression, frantar2022optimal, yuan2023rptq, frantar2023gptq} \textit{CrossGET} is orthogonal to these techniques and does not seek to quantitatively surpass them. Instead, being orthogonal indicates that these techniques can be used together with \textit{CrossGET} to further enhance their acceleration effect. Besides, \textit{CrossGET} offers distinct advantages, including 1) Unlike knowledge distillation that necessitates tuning, \textit{CrossGET} offers the flexibility to be utilized both with and without tuning. This is particularly beneficial when tuning large models is costly or when data are publicly unavailable. 2) The effectiveness of model pruning is heavily dependent on granularity. Unstructured and semi-structured pruning hardly delivers practical speedup without special hardware support, which is unnecessary for \textit{CrossGET}. 3) Low-bit quantization may result in unstable training and necessitate custom CUDA kernel implementations, which are unnecessary for \textit{CrossGET}. Furthermore, a recent advance, TRIPS \cite{jiang2022trips}, employs text feature extracted from the Bert~\cite{devlin2018bert} encoder to unidirectionally guide the token reduction in image encoder, which is limited to modality-independent models. In contrast, \textit{CrossGET} is not only applicable to both modality-independent and modality-dependent scenarios, but also executes the modality-dependent token reduction in a more effective bidirectional manner.

\section{Methodology}

Figure \ref{figure overview} demonstrates that \textit{CrossGET} accelerates vision-language Transformers by ensembling tokens. It is inserted into the middle of Self-Attention and FFN layers in both the vision and language branches. To effectively leverage cross-modal information, \textit{CrossGET} proposes \textit{cross-guided matching and ensemble} (Section \ref{Cross-guided matching}). To achieve reliable token-matching results, \textit{CrossGET} utilizes a parallelizable \textit{complete-graph soft matching} algorithm (Section \ref{Complete-graph soft matching}). 

\begin{figure*}[tb]
    \centering
    \captionsetup{font={small}}
    \includegraphics[width=1.0\linewidth]{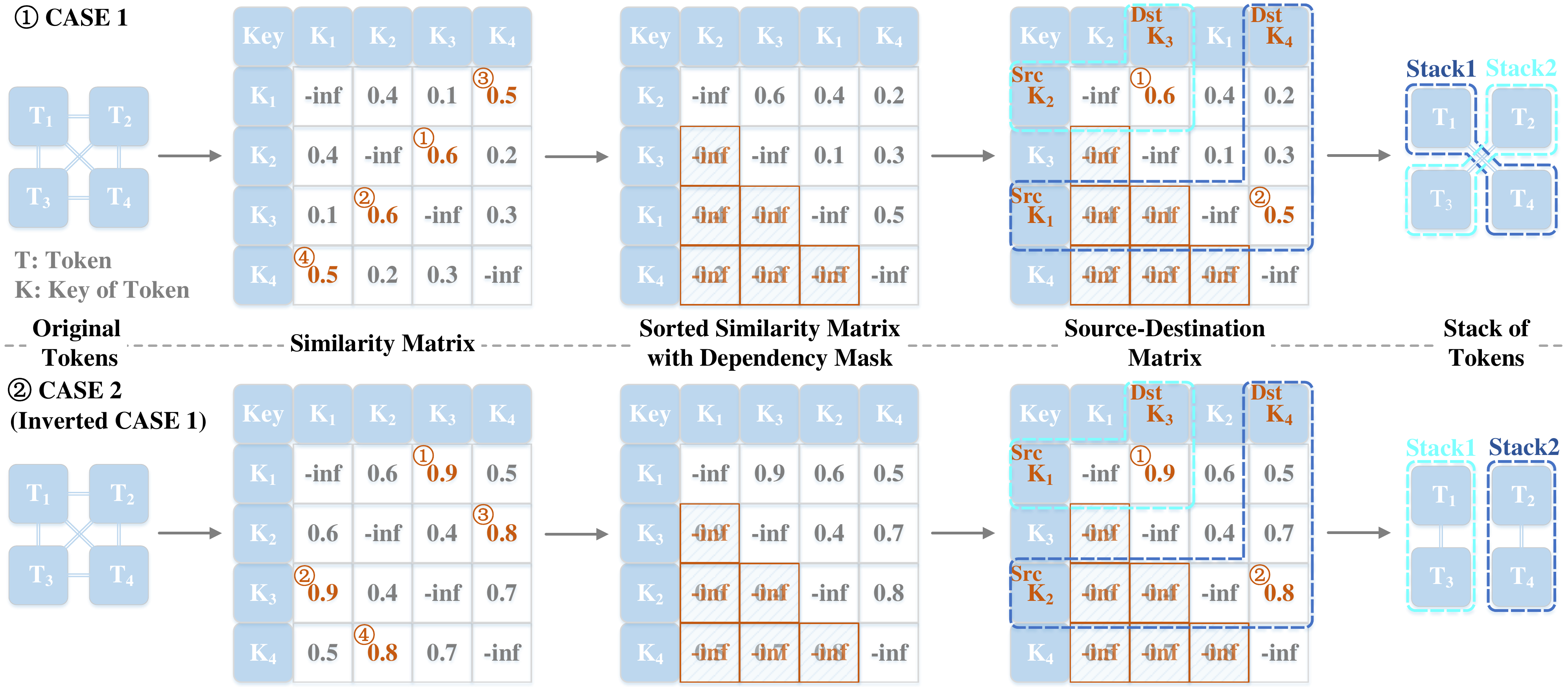}
    \caption{\textbf{Illustration of complete-graph soft matching on two examples.} Case2 is an inverted version of case1 in which the similarity between token pairs in case2 equals ($1 - $ similarity of corresponding pairs in case1).}
    \label{figure match}
    \vspace{-5pt}
\end{figure*}

\subsection{Cross-Guided Matching and Ensemble}
\label{Cross-guided matching}

\paragraph{Dependencies of Calculation Order} \ For multimodal models, in addition to utilizing intra-modal similarity as guidance, token-matching results can further benefit from cross-modal guidance. However, effectively introducing cross-modal guidance is challenging, particularly when dependencies exist on the calculation order of modalities. 

For example, if modality $\mathbb{A}$ requires guidance from modality $\mathbb{B}$, then $\mathbb{B}$ should perform inference, output features as cross-modal guidance, and send these to $\mathbb{A}$. However, if a calculation dependency exists (\eg, the output of $\mathbb{A}$ is a necessary input for $\mathbb{B}$), $\mathbb{B}$ cannot initiate inference before $\mathbb{A}$ completes its inference process. Therefore, $\mathbb{A}$ cannot leverage the cross-modal guidance provided by $\mathbb{B}$.

\paragraph{Breaking Dependencies} \ To allow $\mathbb{A}$ to leverage information from the succeeding modality $\mathbb{B}$ without being constrained by order of calculations, \textit{CrossGET} decouples the capability to guide preceding modalities from the inference process on succeeding modalities, \ie, $\mathbb{B}$ can offer guidance to $\mathbb{A}$ before $\mathbb{B}$'s inference. As illustrated in Figure \ref{figure cross}, this is achieved by injecting learnable cross tokens into each modality, driving them to learn cross-modal information from each other. When conducting inference within a modality, cross tokens act as proxies for other modalities, offering cross-modal guidance on behalf of other modalities.

\paragraph{Cross-Guided Matching} \ Cross tokens provide cross-modal importance $\bm{I}$ as a metric to guide \textit{complete-graph soft matching}. $\bm{I}$ is calculated as the cosine similarity between the query of the cross-token $\bm{T}_c \in \mathbb{R}^{1 \times d}$ where $d$ is the embedding size and the key of other tokens $\bm{T}_i \in \mathbb{R}^{1 \times d} , \, i \neq c$:
\begin{equation}
\label{cross-modal importance}
    \bm{I}_i = \frac{(\bm{T}_c\bm{W}^q)(\bm{T}_i\bm{W}^k)^\top}{\Vert \bm{T}_c\bm{W}^q \Vert_2 \Vert \bm{T}_i\bm{W}^k \Vert_2},
\end{equation}
where $\bm{W}^q, \bm{W}^k \in \mathbb{R}^{d \times d}$ are weights of query and key layers, respectively. $\Vert\cdot\Vert_2$ denotes L2-norm. 

\paragraph{Cross-Guided Ensemble} \ \textit{CrossGET} can be further enhanced by incorporating cross-modal guidance into the ensemble process. More specifically, employing the softmax value of cross-modal importance to produce a weighted summation of the stacked tokens as the ensemble results:
\begin{equation}
    \bm{T}_i = \sum\nolimits_{\bm{T}_j \in \bm{S}_i} \operatorname{softmax}(\bm{I})_j \bm{T}_j,
\end{equation}
where $\bm{S}_i$ represents the set of the stacked tokens, and $\bm{T}_i$ signifies the corresponding ensembled token.

\paragraph{Loss Function} \ JS divergence $\mathcal{L}_\mathcal{JS}$ (\ie, a symmetrized KL divergence $\mathcal{L}_\mathcal{KL}$) between after-projection cross tokens $T^i_{cv}$ from vision and $T^i_{cl}$ from language in layer $i$ is \footnote{For modalities with different number of layers, order-preserving mappings between layer indices can be employed.}:
\begin{gather}
    \mathcal{L}^i_\mathcal{JS} = \mathcal{L}_\mathcal{JS}[(\bm{T}^i_{cv}\tilde{\bm{W}}^{v})||(\bm{T}^i_{cl}\tilde{\bm{W}}^{l})] \\
    = \frac{1}{2}\left[\mathcal{L}_\mathcal{KL}[(\bm{T}^i_{cv}\tilde{\bm{W}}^{v})||\bm{T}^i_m] + \mathcal{L}_\mathcal{KL}[(\bm{T}^i_{cl}\tilde{\bm{W}}^{l})||\bm{T}^i_m]\right], \\
    \bm{T}^i_m = \frac{1}{2}(\bm{T}^i_{cv}\tilde{\bm{W}}^{v} + \bm{T}^i_{cl}\tilde{\bm{W}}^{l}),
\end{gather}
where $\tilde{\bm{W}}^{v}$ and $\tilde{\bm{W}}^{l}$ represent the detached weights of the existing projection layers used for alignment in vision modality and language modality, respectively. Being detached implies that $\mathcal{L}^i_\mathcal{JS}$ produce gradients solely with respect to cross tokens $\bm{T}^i_{cv}$ and $\bm{T}^i_{cl}$, not affecting the projection layers. The weight of projection layers $\bm{W}^v$ and $\bm{W}^l$ are updated exclusively based on the gradients from the original loss. $\mathcal{L}^i_\mathcal{JS}$ is introduced to encourage cross tokens to learn cross-modal information from different modalities:
\begin{equation}
 \mathcal{L} = \mathcal{L_{O}} + \alpha \sum\nolimits_{i=0}^{L-1}\mathcal{L}^i_\mathcal{JS},
\end{equation}
where $\mathcal{L_{O}}$ denotes the original loss for learning a multimodal model, $\alpha$ is a hyperparameter to align the loss items closely in order of magnitude, and $L$ is the number of model layers, which means cross tokens are inserted into each layer of the model and $\mathcal{L_{JS}}$ should be calculated for each layer.

\begin{table*}[t]
  \setlength{\tabcolsep}{0.132cm}
  \captionsetup{font={small}}
  \small
  \centering
  \caption{\textbf{Accelerate CLIP on the Flickr30K dataset of the Image-Text Retrieval task.} R: Recall. R@1, R@5 and R@10 are the higher the better. Experimental results are reported \textit{after training for all approaches}. CrossGET\textsuperscript{\tri} only uses \textit{complete-graph soft matching} (CGSM) (Section \ref{Complete-graph soft matching}), CrossGET\textsuperscript{\dia} adds \textit{cross-guided matching} (CGM) (Section \ref{Cross-guided matching}) on \textsuperscript{\tri}, and CrossGET\textsuperscript{\hollowstar} further adds \textit{cross-guided ensemble} (CGE) (Section \ref{Cross-guided matching}) on \textsuperscript{\dia}. Here \textit{UPop} uses a larger CLIP as its original model, and therefore GFLOPs is higher.}
  \vspace{-3pt}
  \begin{tabular}{ l @{\hspace{2\tabcolsep}} l l l l l l l l l}
    \toprule
    \multirow{2}{*}[-3pt]{Approach} & \multicolumn{3}{c}{Image $\rightarrow$ Text} & \multicolumn{3}{c}{Text $\rightarrow$ Image} & \multicolumn{1}{c}{Avg.} &\multirow{2}{*}[-3pt]{\makecell{GFLOPs \\ $\downarrow$}} & \multirow{2}{*}[-3pt]{\makecell{Throughput \\ $\uparrow$}} \\ 
    \cmidrule(r){2-4} \cmidrule(lr){5-7} \cmidrule(r){8-8}
    & R@1 & R@5 & R@10 & R@1 & R@5 & R@10 & $\overline{\textbf{R@1}}$ \\
    \cmidrule{1-10}
    CLIP \cite{radford2021learning} & 92.1 & \fade{99.1} & \fade{99.7} & 79.3 & \fade{95.7} & \fade{98.0} & 85.7 & 20.6 & 255.2 \\
    \cmidrule{1-10}
    TRIPS \cite{jiang2022trips} & 90.4 & \fade{98.9} & \fade{99.5} & 76.8 & \fade{94.4} & \fade{97.2} & 83.6 & 16.4 & 316.9 \\
    UPop \cite{pmlr-v202-shi23e}  & 82.9 & \fade{95.7} & \fade{97.8} & 67.3 & \fade{89.5} & \fade{93.5} & 75.1 & 51.3 & - \\
    \cmidrule{1-10}
    Hourglass \cite{liang2022expediting} & 90.5 & \fade{99.0} & \fade{99.7} & 77.9 & \fade{94.8} & \fade{97.3} & 84.2 & 15.0 & 342.3 \\
    DynamicViT \cite{rao2021dynamicvit} & 89.4 & \fade{98.8} & \fade{99.3} & 75.7 & \fade{94.2} & \fade{97.0} & 82.6 & 12.2 & 422.1 \\
    EViT \cite{liang2022not} & 89.9 & \fade{98.6} & \fade{99.4} & 76.7 & \fade{94.5} & \fade{97.4} & 83.3 & 12.4 & 413.2 \\
    ToMe \cite{bolya2022tome}  & 90.8$_{\color{red}\downarrow 1.3}$ & \fade{99.2}$_{\color{ForestGreen}\uparrow 0.1}$ & \fade{99.5}$_{\color{red}\downarrow 0.2}$ & 78.1$_{\color{red}\downarrow 1.2}$ & \fade{95.3}$_{\color{red}\downarrow 0.4}$ & \fade{97.7}$_{\color{red}\downarrow 0.3}$ & 84.5$_{\color{red}\downarrow 1.2}$ & 11.8  & 417.4 \\
    ToMe+Extra Token & 90.8$_{\color{red}\downarrow 1.3}$ & \fade{98.7}$_{\color{red}\downarrow 0.4}$ & \fade{99.6}$_{\color{red}\downarrow 0.1}$ & 78.8$_{\color{red}\downarrow 0.5}$ & \fade{95.1}$_{\color{red}\downarrow 0.6}$ & \fade{97.6}$_{\color{red}\downarrow 0.4}$ & 84.8$_{\color{red}\downarrow 0.9}$ & 11.9 & 412.9 \\
    ToMe+CGM$\&$CGE & 91.5$_{\color{red}\downarrow 0.6}$ & \fade{99.0}$_{\color{red}\downarrow 0.1}$ & \fade{99.6}$_{\color{red}\downarrow 0.1}$ & 78.6$_{\color{red}\downarrow 0.7}$ & \fade{95.4}$_{\color{red}\downarrow 0.3}$ & \fade{97.8}$_{\color{red}\downarrow 0.2}$ & 85.1$_{\color{red}\downarrow 0.6}$ & 11.9 & 409.9 \\
    \cmidrule{1-10}
    \rowcolor{gray!10}CrossGET\textsuperscript{\tri} (CGSM) & 90.9$_{\color{red}\downarrow 1.2}$ & \fade{99.2}$_{\color{ForestGreen}\uparrow 0.1}$ & \textbf{\fade{99.9}}$_{\color{ForestGreen}\uparrow 0.2}$ & 79.1$_{\color{red}\downarrow 0.2}$ & \fade{95.1}$_{\color{red}\downarrow 0.6}$ & \fade{97.6}$_{\color{red}\downarrow 0.4}$ & 85.0$_{\color{red}\downarrow 0.7}$ & 11.9 & 408.9 \\
    \rowcolor{gray!25}CrossGET\textsuperscript{\dia} (CGSM+CGM) & \textbf{92.1}$_{\color{ForestGreen}\uparrow 0.0}$ & \fade{99.3}$_{\color{ForestGreen}\uparrow 0.2}$ & \fade{99.7}$_{\color{ForestGreen}\uparrow 0.0}$ & 79.5$_{\color{ForestGreen}\uparrow 0.2}$ & \fade{95.3}$_{\color{red}\downarrow 0.4}$ & \fade{97.7}$_{\color{red}\downarrow 0.3}$ & 85.8$_{\color{ForestGreen}\uparrow 0.1}$ & 12.0 & 402.1 \\
    \rowcolor{gray!35}CrossGET\textsuperscript{\hollowstar} (CGSM+CGM$\&$CGE) & \textbf{92.1}$_{\color{ForestGreen}\uparrow 0.0}$ & \textbf{\fade{99.7}}$_{\color{ForestGreen}\uparrow 0.6}$ & \fade{99.8}$_{\color{ForestGreen}\uparrow 0.1}$ & \textbf{79.6}$_{\color{ForestGreen}\uparrow 0.3}$ & \textbf{\fade{95.7}}$_{\color{ForestGreen}\uparrow 0.0}$ & \textbf{\fade{98.0}}$_{\color{ForestGreen}\uparrow 0.0}$ & \textbf{85.9}$_{\color{ForestGreen}\uparrow 0.2}$ & 12.0$_{\color{ForestGreen}\downarrow 42\%}$ & 401.8$_{\color{ForestGreen}\uparrow 57\%}$\\
    \bottomrule
  \end{tabular}
  \vspace{-8pt}
  \label{table CLIP Retrieval}
\end{table*}

\subsection{Complete-Graph Soft Matching}
\label{Complete-graph soft matching}
\paragraph{Problem Formulation for Token Matching} Token matching is aimed at determining which tokens should be combined. Suppose there are $N \in \mathbb{N}^{+}$ tokens in total, and $r \in \mathbb{N}^{+} $ ($r < N$) tokens among them should be eliminated (\ie, combined together with other tokens), then the token matching problem can be formulated as a discrete optimization problem that is to find a set of feasible token pairs: 
\begin{equation}
    \bm{P} = \{(\bm{T}_i, \bm{T}_j) \, | \, 0 \leq i,j \leq N, i \neq j\}, \quad |\bm{P}| = r,
\end{equation}
where $\bm{T}_i$ denotes tokens $i$, and $|\cdot|$ denotes the size of the set, to maximize the objective function
\begin{equation}
    S = \sum\nolimits_{(\bm{T}_i, \bm{T}_j) \in P} \mathcal{D}(\bm{T}_i, \bm{T}_j),
\end{equation} 
where $\mathcal{D}$ is a function (\eg, cosine similarity) that calculates the similarity between the key of the token $\bm{T}_i$ and $\bm{T}_j$. Appendix \ref{section example token matching} provides examples to elaborate.

\paragraph{Parallelizability} While iterative clustering can be utilized for token matching, it cannot be parallelized and is time-consuming. To facilitate the parallelizability, an additional constraint $\bm{T}^S \cap \bm{T}^D = \phi,$ should be met.
\ie, the source set $\bm{T}^S$ and destination set $\bm{T}^D$ should be disjointed, where
\begin{align}
    \bm{T}^S & = \{\bm{T}_i \, | \, (\bm{T}_i, \bm{T}_j) \in P\}, \quad |\bm{T}^S| = r, \\
    \bm{T}^D & = \{\bm{T}_j \, | \, (\bm{T}_i, \bm{T}_j) \in P\}, \quad |\bm{T}^D| \leq r.
\end{align}
\paragraph{Algorithm Procedure} \textit{Complete-Graph Soft Matching} is designed as a non-iterative, approximate algorithm to ensure parallelizability and high efficiency. It enables each token to consider its similarity with all other tokens, as shown in Figure \ref{figure match} (Appendix \ref{section algorithm cgsm} provides an implementation):

\begin{itemize}
\item \textbf{Step 1}: Calculate the cosine similarities $\frac{\bm{K}\bm{K}^{\top}}{\Vert \bm{K} \Vert_2^2}$ between the keys $\bm{K}$ of every two tokens to generate the similarity matrix $\bm{D} \in \mathbb{R}^{N \times N}$ (Diagonal self-similarities are ignored). 

\item \textbf{Step 2}: Sort the rows and columns of the similarity matrix in descending order based on their maximum similarity $\max \limits_{1 \leq j \leq N}\bm{D}_{ij}$ and $\max \limits_{1 \leq i \leq N}\bm{D}_{ij}$ to other tokens.

\item \textbf{Step 3}: Upon the sorted similarity matrix $\bm{D}^{\star}$, a lower triangle dependency mask $\bm{M}_{ij}= \begin{cases}-\infty & \text { for } i \geq j \\ 0 & \text { for } i < j\end{cases}$ is applied to disjoint the sets $\bm{T}^S$ and $\bm{T}^D$. It explicitly prioritizes the matching among tokens based on similarity values, ensuring source tokens with higher priority do not become targets for those with lower priority.

\item \textbf{Step 4}: Select $r$ rows with the highest similarity $\max \limits_{1 \leq j \leq N}\bm{D}^{\star}_{ij}$ to other tokens as the source set $\bm{T}^S$. For every token in $\bm{T}^S$, select tokens from $\bm{T} \backslash \bm{T}^S$ that exhibit the highest similarity as the destination set $\bm{T}^D$. 

\item \textbf{Step 5}: The matching among tokens leads to multiple connected components (\ie, stacks), and tokens in each stack are ensembled by averaging. 
\end{itemize}

This procedure is non-iterative and parallelizable. As depicted in Figure \ref{figure match}, \textit{complete-graph soft matching} achieves optimal solutions in both case1 and case2. 

\paragraph{Incorporation with Cross-Guided Matching and Ensemble} Modifications are as follows to leverage cross-modal guidance (Appendix \ref{section algorithm cgem} provides an implementation):

$\bullet$ \textbf{Step 4}: Select $r$ rows via metric $\max \limits_{1 \leq j \leq N}\bm{D}^{\star}_{ij} - \bm{I}$ (\ie, highest similarity to other tokens - cross-modal importance) instead of  $\max \limits_{1 \leq j \leq N}\bm{D}^{\star}_{ij}$.

$\bullet$ \textbf{Step 5}: Instead of averaging, ensembling tokens via weighted summation based on $\operatorname{softmax}(\bm{I})$. 

Appendix \ref{section sub optimal} and \ref{section expectation} provide additional discussions on the sub-optimal cases of this method and analyses regarding the expectation of optimal matching probability, respectively.

\begin{table}[t]
\captionsetup{font={small}}
\setlength{\tabcolsep}{0.1mm}
\captionsetup{font={small}}
\small
\centering
\caption{\textbf{Accelerate BLIP on the NLVR2 dataset of the Vision Reasoning task.} BLIP is the original model for all approaches.}
    \begin{tabular}{l l l l l}
    \toprule
    Approach  & Dev Acc & Test Acc & GFLOPs & Throughput \\
    \cmidrule(r){1-1} \cmidrule(r){2-3} \cmidrule{4-5}
    BLIP \cite{li2022blip} & \fade{82.3} & 83.4 & 132.5 & 39.8 \\
    \cmidrule(r){1-1} \cmidrule(r){2-3} \cmidrule{4-5}
    UPop \cite{pmlr-v202-shi23e} & \fade{80.3}$_{\color{red}\downarrow 2.0}$ & 81.1$_{\color{red}\downarrow 2.3}$ & 89.4 & - \\
    ToMe \cite{bolya2022tome} & \fade{81.7}$_{\color{red}\downarrow 0.6}$ & 82.2$_{\color{red}\downarrow 1.2}$ & 59.0 & 81.9 \\
    \rowcolor{gray!10}CrossGET\textsuperscript{\tri} (CGSM) & \textbf{\fade{82.2}}$_{\color{red}\downarrow 0.1}$ & 
    82.6$_{\color{red}\downarrow 0.8}$ & 60.8 & 77.7 \\
    \rowcolor{gray!25}CrossGET\textsuperscript{\hollowstar} (Ours) & \fade{82.1}$_{\color{red}\downarrow 0.2}$ & \textbf{83.2}$_{\color{red}\downarrow 0.2}$ & 61.1$_{\color{ForestGreen}\downarrow 57\%}$ & 76.8$_{\color{ForestGreen}\uparrow 93\%}$ \\
    \bottomrule
  \end{tabular}
\label{table nlvr}
\end{table} 

\begin{table*}[t]
  \setlength{\tabcolsep}{2mm}
  \captionsetup{font={small}}
  \small
  \centering
  \caption{\textbf{Accelerate BLIP on the COCO Caption dataset of the Image Caption task.} The suffix -F denotes GFLOPs and throughput for the forward, while -G denotes GFLOPs and throughput for the generation.}
  \begin{tabular}{l l l l l l l}
    \toprule
    Approach  & CIDEr & SPICE & GFLOPs-F & Throughput-F & GFLOPs-G & Throughput-G\\
    \cmidrule(r){1-1} \cmidrule(r){2-3} \cmidrule(r){4-5} \cmidrule{6-7}
    BLIP \cite{li2022blip} & 133.3 & 23.8 & 65.7 & 106.4 & 330.7 & 17.2 \\
    \cmidrule(r){1-1} \cmidrule(r){2-3} \cmidrule(r){4-5} \cmidrule{6-7}
    UPop \cite{pmlr-v202-shi23e} & 128.9$_{\color{red}\downarrow 4.4}$ & 23.3$_{\color{red}\downarrow 0.5}$ & 39.8 & - & - & - \\
    ToMe \cite{bolya2022tome} & 130.3$_{\color{red}\downarrow 3.0}$ & 23.3$_{\color{red}\downarrow 0.5}$ & 29.2 & 209.3 & 43.8 & 77.7 \\
    \rowcolor{gray!10}CrossGET (Ours) & \textbf{131.6}$_{\color{red}\downarrow 1.7}$ & \textbf{23.8}$_{\color{ForestGreen}\uparrow 0.0}$ & 30.1$_{\color{ForestGreen}\downarrow 54\%}$ & 183.5$_{\color{ForestGreen}\uparrow 72\%}$ & 46.7$_{\color{ForestGreen}\downarrow 86\%}$ & 73.9$_{\color{ForestGreen}\uparrow 330\%}$ \\
    \bottomrule
  \end{tabular}
  \label{table caption coco}
\end{table*}

\begin{figure*}[t]
    \captionsetup{font={small}}
    \centering
    \includegraphics[width=0.85\linewidth]{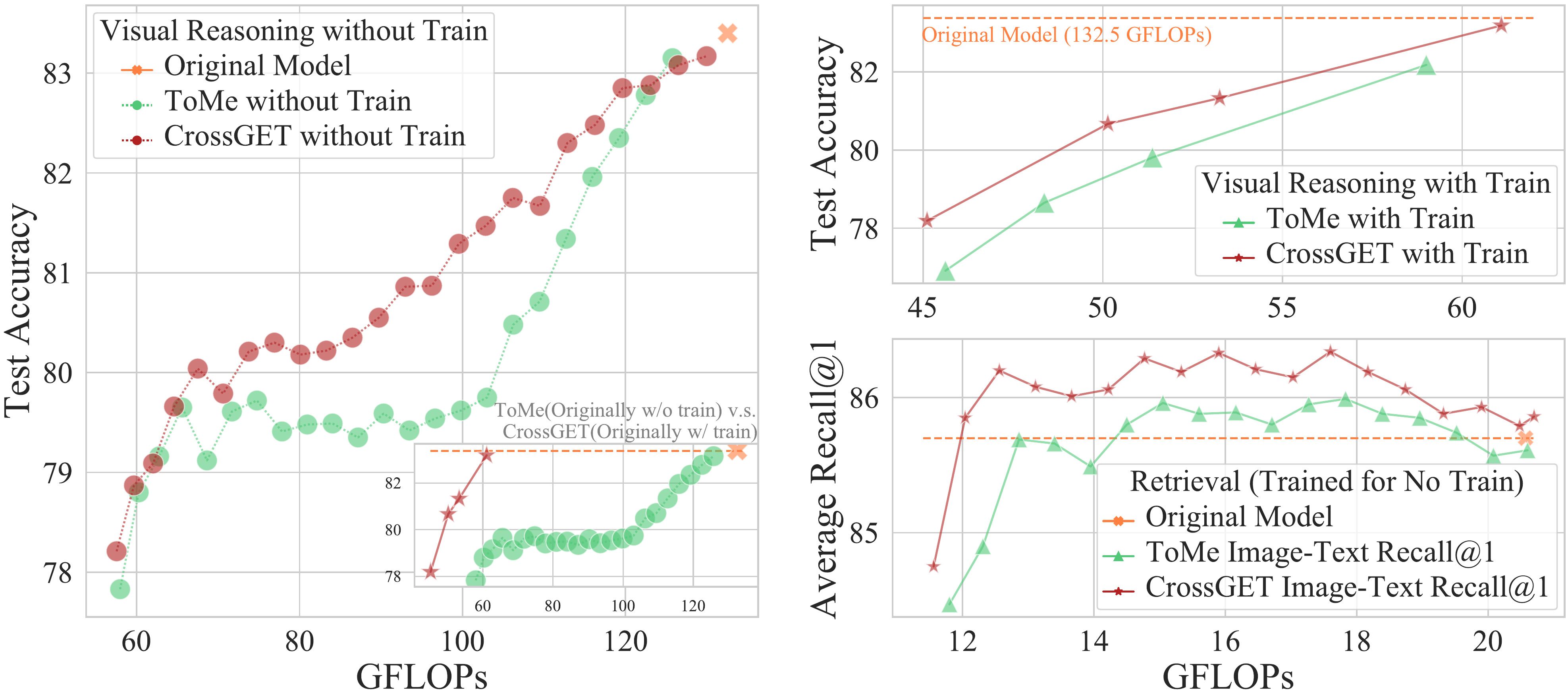}
    \caption{\textbf{Performance-Cost tradeoffs in three situations:} 1) The left subfigure illustrates the tradeoff for BLIP on the NVLR2 dataset of the Visual Reasoning task without training. 2) The upper-right subfigure illustrates the tradeoff for BLIP on the NVLR2 dataset of the Visual Reasoning task with training. 3) The lower-right subfigure illustrates the tradeoff for CLIP on the Flickr30K dataset of the Image-Text Retrieval task are trained with $50\%$ token reduced and then re-evaluated under other token reduction ratios without training.}
    \label{figure curves}
    \vspace{-6pt}
\end{figure*}

\begin{table*}[!h]
 \setlength{\tabcolsep}{1.8mm}
  \captionsetup{font={small}}
  \small
  \centering
  \caption{\textbf{Accelerate BLIP on the NoCaps dataset of the Novel Object Caption task.} All metrics are the higher the better, and the evaluation uses the same model finetuned on the COCO Caption dataset as in Table \ref{table caption coco}, and therefore the GFLOPs and throughput of models are the same as in Table \ref{table caption coco}.}
  \begin{tabular}{ l l l l l l l l l }
    \toprule
    \multirow{2}{*}[-2pt]{Approach} & \multicolumn{2}{c}{in-domain} & \multicolumn{2}{c}{near-domain} & \multicolumn{2}{c}{out-domain} & \multicolumn{2}{c}{\textbf{entire}} \\ 
    \cmidrule(r){2-3} \cmidrule(r){4-5} \cmidrule(r){6-7} \cmidrule{8-9}
    & CIDEr & SPICE & CIDEr & SPICE & CIDEr & SPICE & CIDEr & SPICE  \\
    \cmidrule(r){1-1} \cmidrule(r){2-3} \cmidrule(r){4-5} \cmidrule(r){6-7} \cmidrule{8-9}
    BLIP \cite{li2022blip} & \fade{111.9} & \fade{14.9} & \fade{108.8} & \fade{14.8} & \fade{112.1} & \fade{14.2} & 109.9 & 14.7 \\
    \cmidrule(r){1-1} \cmidrule(r){2-3} \cmidrule(r){4-5} \cmidrule(r){6-7} \cmidrule{8-9}
    ToMe \cite{bolya2022tome} & \fade{107.9}$_{\color{red}\downarrow 4.0}$ & \fade{14.8}$_{\color{red}\downarrow 0.1}$ & \fade{105.1}$_{\color{red}\downarrow 3.7}$ & \fade{14.4}$_{\color{red}\downarrow 0.4}$ & \fade{106.4}$_{\color{red}\downarrow 5.7}$ & \textbf{\fade{14.1}}$_{\color{red}\downarrow 0.1}$ & 105.7$_{\color{red}\downarrow 4.2}$ & 14.4$_{\color{red}\downarrow 0.3}$ \\
    \rowcolor{gray!10}CrossGET (Ours) & \textbf{\fade{113.2}}$_{\color{ForestGreen}\uparrow 1.3}$ & \textbf{\fade{15.1}}$_{\color{ForestGreen}\uparrow 0.2}$ & \textbf{\fade{107.2}}$_{\color{red}\downarrow 1.6}$ & \textbf{\fade{14.6}}$_{\color{red}\downarrow 0.2}$ & \textbf{\fade{107.4}}$_{\color{red}\downarrow 4.7}$ & \textbf{\fade{14.1}}$_{\color{red}\downarrow 0.1}$ & \textbf{108.1}$_{\color{red}\downarrow 1.8}$ & \textbf{14.6}$_{\color{red}\downarrow 0.1}$ \\
    \bottomrule
  \end{tabular}
  \label{table caption Nocaps }
\end{table*}

\begin{table*}[t]
  \setlength{\tabcolsep}{1.8mm}
  \captionsetup{font={small}}
  \small
  \centering
  \caption{\textbf{Accelerate BLIP on the VQA2.0 dataset of the Visual Question Answer task.} "yes/no", "number", "other", and "overall" denote accuracy on the corresponding types of questions. These four metrics are the higher the better. The suffix -F denotes GFLOPs and throughput for the forward that a single image may be accompanied by multiple questions and answers during training, while -T denotes GFLOPs and throughput for the test that a single image is accompanied by only one question and answer.}
  \begin{tabular}{l l l l l l l l l }
    \toprule
    Approach  & yes/no & number & other & \textbf{overall} & GFLOPs-F & Throughput-F & GFLOPs-T & Throughput-T \\
    \cmidrule(r){1-1} \cmidrule(r){2-4} \cmidrule(r){5-5} \cmidrule(r){6-7} \cmidrule{8-9}
    BLIP \cite{li2022blip} & \fade{92.6} & \fade{60.6} & \fade{68.3} & 77.4 & 186.1 & 67.2 & 106.8 & 53.0 \\
    \cmidrule(r){1-1} \cmidrule(r){2-4} \cmidrule(r){5-5} \cmidrule(r){6-7} \cmidrule{8-9}
    UPop \cite{pmlr-v202-shi23e} & \fade{-} & \fade{-} & \fade{-} & 76.3$_{\color{red}\downarrow 1.1}$ & 109.4 & - & - & - \\
    ToMe \cite{bolya2022tome} & \fade{92.1}$_{\color{red}\downarrow 0.5}$ & \fade{59.3}$_{\color{red}\downarrow 1.3}$ & \fade{67.1}$_{\color{red}\downarrow 1.2}$ & 76.5$_{\color{red}\downarrow 0.9}$ & 119.0 & 141.1 & 46.7 & 90.1 \\
    \rowcolor{gray!10}CrossGET (Ours) & \textbf{\fade{92.4}}$_{\color{red}\downarrow 0.2}$ & \textbf{\fade{59.7}}$_{\color{red}\downarrow 0.9}$ & \textbf{\fade{67.7}}$_{\color{red}\downarrow 0.6}$ & \textbf{77.0}$_{\color{red}\downarrow 0.4}$ & 124.5$_{\color{ForestGreen}\downarrow 33\%}$ & 120.4$_{\color{ForestGreen}\uparrow 79\%}$ & 49.0$_{\color{ForestGreen}\downarrow 54\%}$ & 81.3$_{\color{ForestGreen}\uparrow 53\%}$ \\
    \bottomrule
  \end{tabular}
  \label{table vqa}
\end{table*}

\begin{table*}[t]
 \setlength{\tabcolsep}{1.3mm}
  \captionsetup{font={small}}
  \small
  \centering
  \caption{\textbf{Accelerate multimodal LLM BLIP2-OPT6.7B on the COCO Caption dataset of the Image Caption task.} The suffix -F denotes GFLOPs and throughput for the forward, while -G denotes GFLOPs and throughput for the generation. \textsuperscript{$*$} indicates using greedy decoding instead of beam search for generation. Experimental results on BLIP2-OPT2.7B are provided in Appendix \ref{section opt}. }
  \begin{tabular}{l l l l l l l l l}
    \toprule
    Approach & Tuning & CIDEr & BLEU@4 & GFLOPs-F & Throughput-F & GFLOPs-G & Throughput-G & Throughput-G\textsuperscript{$*$}  \\
    \cmidrule(r){1-2} \cmidrule(r){3-4} \cmidrule(r){5-6} \cmidrule{7-9}
    BLIP2-OPT6.7B & - & 144.5 & 42.5 & 1042.6 & 47.4 & 2461.1 & 16.2 & 46.2 \\
    \cmidrule(r){1-2} \cmidrule(r){3-4} \cmidrule(r){5-6} \cmidrule{7-9}
    & \fade{w/o tuning} & \fade{144.7} & \textbf{\fade{42.4}} & \fade{957.6} & \fade{-} & \fade{2342.7} & \fade{-} & \fade{-} \\
    & \fade{w/o tuning} & \fade{144.3} & \fade{42.3} & \fade{868.1} & \fade{-} & \fade{2086.7} & \fade{-} & \fade{-} \\
    & \fade{w/o tuning} & \fade{142.4} & \fade{41.9} & \fade{780.7} & \fade{-} & \fade{2232.4} & \fade{-} & \fade{-} \\
    & \fade{w/o tuning} & \fade{135.5} & \fade{40.1} & \fade{695.1} & \fade{-} & \fade{2046.9} & \fade{-} & \fade{-} \\
    \cmidrule(r){2-2} \cmidrule(r){3-4} \cmidrule(r){5-6} \cmidrule{7-9}
    \multirow{-5}{*}{\makecell[l]{ToMe \\ \cite{bolya2022tome}}} & w/ tuning & 141.7$_{\color{red}\downarrow 2.8}$ & 41.4$_{\color{red}\downarrow 1.1}$ & 544.8 & 92.6 & 1510.0 & 21.5 & 75.1 \\
    \midrule
    & \fade{w/o tuning} & \fade{144.5} & \fade{42.3} & \fade{973.8} & \fade{-} & \fade{2392.3} & \fade{-} & \fade{-} \\
    & \fade{w/o tuning} & \fade{144.6} & \textbf{\fade{42.4}} & \fade{881.1} & \fade{-} & \fade{2266.2} & \fade{-} & \fade{-} \\
    & \fade{w/o tuning} & \fade{143.3} & \textbf{\fade{42.1}} & \fade{790.9} & \fade{-} & \fade{2176.1} & \fade{-} & \fade{-} \\
    & \fade{w/o tuning} & \fade{137.5} & \textbf{\fade{40.6}} & \fade{703.4} & \fade{-} & \fade{2121.8} & \fade{-} & \fade{-} \\
    \cmidrule(r){2-2} \cmidrule(r){3-4} \cmidrule(r){5-6} \cmidrule{7-9}
    \multirow{-5}{*}{\makecell{CrossGET(Ours)}} & \cellcolor{gray!10}w/ tuning & \cellcolor{gray!10}\textbf{143.1}$_{\color{red}\downarrow 1.4}$ & \cellcolor{gray!10}\textbf{42.0}$_{\color{red}\downarrow 0.5}$ & \cellcolor{gray!10}558.2$_{\color{ForestGreen}\downarrow 49\%}$ & \cellcolor{gray!10}91.0$_{\color{ForestGreen}\uparrow 92\%}$ & \cellcolor{gray!10}1583.2$_{\color{ForestGreen}\downarrow 36\%}$ & \cellcolor{gray!10}21.6$_{\color{ForestGreen}\uparrow 33\%}$ & \cellcolor{gray!10}75.7$_{\color{ForestGreen}\uparrow 64\%}$ \\
    \bottomrule
  \end{tabular}
  \label{table blip2 large caption}
\end{table*}
\section{Experiments}

We report the performance on modality-independent model CLIP \cite{radford2021learning} as well as modality-dependent models BLIP/BLIP2 \cite{li2022blip, li2023blip}, and mainstream tasks such as Image-Text Retrieval, Visual Reasoning, Image Captioning, and Visual Question Answering.

\subsection{Experiments with CLIP on Image-Text Retrieval}
We conduct experiments on the CLIP model, and Flickr30K datasets \cite{young2014image} with Karpathy split \cite{karpathy2015deep} of Image-Text Retrieval and Text-Image Retrieval task. The number of tokens is reduced to half with the same reduction number for each layer. For example, suppose one of the modalities of a 12-layer CLIP has 100 tokens as input, then $\lfloor\frac{100}{12} \rfloor=8$ tokens will be eliminated from each layer so that the number of tokens left in the last layer is $100-12\times 8=4$, and the total number of tokens across all layers is roughly reduced to half. If not specified, the number of tokens to be reduced in other experiments is also determined by this strategy. 

\paragraph{Comparison with Baselines} Unless stated otherwise, all reported experimental results are after training. Table \ref{table CLIP Retrieval} demonstrates that \textit{CrossGET} outperforms both the SOTA multimodal model pruning approach UPop \cite{pmlr-v202-shi23e}, token reduction approach TRIPS \cite{jiang2022trips}, and other unimodal token reduction approachs \cite{bolya2022tome, liang2022not, rao2021dynamicvit, liang2022expediting} without extra learnable parameter other than negligible cross tokens \footnote{For fairness of comparison, methods that require additional learnable parameters exceeding the level of several tokens are not taken into comparison (\eg, simply adding a new linear projection layer with weight $W \in \mathbb{R}^{768\times 768}$ already needs 768 times the number of our cross token's parameters $T_c\in \mathbb{R}^{1\times 768}$)}. 
It can also be observed that simply adding an extra learnable token to unimodal approach ToMe does not bring a notable improvement. In particular, the average of Recall@1 is significantly lower than \textit{CrossGET}, which indicates that the improvement given by \textit{cross-guided matching and ensemble} is mainly from learning cross-modal information instead of the increase of learnable tokens.

\paragraph{Effect of individual components} As highlighted by grey in Table~\ref{table CLIP Retrieval}, \textit{complete-graph soft matching} (CGSM) brings improvements on most of the metrics and a significant improvement on text-to-image retrieval (recall@1 increases from 78.1 to 79.1). Since the complete graph has more similarity of token pairs to compute than the bipartite graph, GFLOPs also slightly increase by 0.1. \textit{Cross-guided matching} (CGM) brings further improvement on most metrics and a significant improvement on image-to-text retrieval (recall@1 increases from 90.9 to 92.1). Since cross tokens interact with other tokens during the forward, GFLOPs again slightly increase by 0.1. \textit{Cross-guided ensemble} brings final improvement on all metrics with negligible extra GFLOPs. Moreover, consistent improvements can also be observed when \textit{Cross-Guided Matching and Ensemble} is applied to \textit{ToMe}. Compared with the original CLIP, \textit{CrossGET} achieves the same image-to-text recall@1 and 0.3 higher text-to-image recall@1 while saving 42$\%$ GFLOPs and improving throughput by 57$\%$. 

\subsection{Experiments with BLIP on Visual Reasoning}

Table \ref{table nlvr} shows \textit{CrossGET} also achieves very competitive performance on the BLIP model and NLVR2 dataset of a vision reasoning task that requires predicting whether a given sentence can describe a pair of given images. Compared with the original BLIP, \textit{CrossGET} gets only 0.2 lower accuracies on the dev set and test set while saving 57$\%$ GFLOPs and improving throughput by 93$\%$. 

\subsection{Experiments at Different Reduction Ratios}

Figure \ref{figure curves} illustrates experimental results at various reduction ratios under three different settings: (1) Comparisons without training (left subfigure). Note that the only part of \textit{CrossGET} that requires training is learning cross tokens. However, they are initialized with informative features (see Appendix \ref{section initialization}) and already contain representative information even though they are not trained. Therefore, \textit{CrossGET} can also be used without training (certainly worse than with training); (2) Comparisons with training (upper-right subfigure). (3) Re-evaluate a trained model (50$\%$ token reduced) under other token reduction ratios without training (lower-right subfigure). These subfigures demonstrate that \textit{CrossGET} achieves superior Pareto frontiers in all three situations. Appendix \ref{section no training}, \ref{section training}, and \ref{section trained for no train} provide detailed data. Besides, the original ToMe method does not require training, and the comparison with it is illustrated in the small plot at the lower right corner of the left subfigure.

\subsection{Experiments with BLIP on Image Captioning}

On auto-regressive models performing cross-modal interactions at each layer and forward via Cross-Attentions, such as the BLIP-Captioning \cite{li2022blip} model, \textit{CrossGET} achieves higher speedups. As shown in Table \ref{table caption coco}, reducing the total tokens by half for the generation brings 86$\%$ saving of GFLOPs and improving 330$\%$ throughput. 

We also conduct experiments on the NoCaps \cite{agrawal2019nocaps} datasets of the Novel Object Caption task, and the model accelerated by \textit{CrossGET} again achieves superior performances on the entire task and all sub-tasks.

\subsection{Experiments with BLIP on Visual QA }

We conduct experiments on the BLIP model \cite{li2022blip} and the test-dev set of the VQA2.0 dataset \cite{goyal2017making}. Table \ref{table vqa} demonstrates that \textit{CrossGET} can also considerably save computational cost and improve throughput for the Visual Question Answering task. For example, when compared with the original model, \textit{CrossGET} gets only 0.4 lower overall accuracy on all three types of questions while saving $33\%$ GFLOPs and improving throughput by $79\%$ for the multiple-question scenario, and saving $54\%$ GFLOPs and improving throughput by $53\%$ for the single-question scenario.

\subsection{Experiments with BLIP2 on Image Captioning}

We apply CrossGET to the multimodal LLM BLIP2 \cite{li2023blip}. Following the original strategy of BLIP2, which tunes the ViT and Q-Former (a BERT) while freezing the LLM, we conduct experiments with and without tuning. Table \ref{table blip2 large caption} demonstrates that CrossGET consistently achieves promising performance on multimodal LLMs. Additionally, compared with BLIP, the language branch of BLIP2 receives fewer tokens from the vision branch, resulting in less generation speedup.

\subsection{Experiments with LLaVA-1.5 on Various Datasets}

For experiments on LLaVA-1.5 \cite{liu2023improved}, we followed its supervised fine-tuning (SFT) strategy, which tunes the LLM (\ie, Vicuna \cite{vicuna2023}) and projector (\ie, MLP) while freezing the ViT. We evenly sampled 10$\%$ of data from the SFT dataset of LLaVA-1.5 as our training dataset. We observed that using more data provided limited improvement in performance recovery for models after acceleration. As shown in Table \ref{table llava}, with only 10$\%$ of the SFT data, CrossGET nearly doubles the throughput of the model forward and improves the throughput of generation by nearly 50$\%$, while maintaining more than 98$\%$ of the original models' capabilities on average. Table \ref{table llava} also indicates that with similar computational cost and throughput, LLaVA-1.5-13B after acceleration achieves better overall performance than LLaVA-1.5-7B without acceleration, which further demonstrates that instead of training smaller models from scratch, CrossGET can efficiently create more capable models from large-scale ones.

\begin{table*}[t]
\small
\setlength{\tabcolsep}{0.45mm}
\captionsetup{font={small}}
\centering
\caption{\textbf{Accelerate multimodal LLM LLaVA-1.5-7B and LLaVA-1.5-13B.} Quantitative evaluation is conducted on ten widely used datasets. Tput represents throughput. The superscript $^\text{F}$ denotes GFLOPs and throughput for the forward, while $^\text{G}$ denotes GFLOPs and throughput for the generation. Details of each dataset are provided in Appendix \ref{section llava benchmark}.}
\begin{tabular}{lccccccccccllll}
\toprule
Approach & VQA$^\text{v2}$ & GQA & VisWiz & SQA$^\text{I}$ & VQA$^\text{T}$ & POPE & MME & MMB & MMB$^\text{CN}$ & SEED$^\text{I}$ & GFLOPs$^\text{F}$ & Tput$^\text{F}$ & GFLOPs$^\text{G}$ & Tput$^\text{G}$ \\
\cmidrule(r){1-11} \cmidrule{12-15}
LLaVA-1.5-7B & 78.5 & 62.0 & 50.0 & 66.8 & 58.2 & 85.9 & 1510.7 & 64.3 & 58.3 & 66.2 & 4480.9 & 32.0 & 6216.7 & 1.7 \\
\rowcolor{gray!15} with CrossGET & 77.3 & 61.4 & 47.7 & 66.7 & 54.9 & 83.9 & 1510.2 & 64.7 & 55.2 & 64.4 & 2382.5$_{\color{ForestGreen}\downarrow 31\%}$ & 60.4$_{\color{ForestGreen}\uparrow 89\%}$ & 4098.4$_{\color{ForestGreen}\downarrow 34\%}$ & 2.5$_{\color{ForestGreen}\uparrow 47\%}$ \\
\midrule
LLaVA-1.5-13B & 80.0 & 63.3 & 53.6 & 71.6 & 61.3 & 85.9 & 1531.3 & 67.7 & 63.6 & 68.2 & 8505.3 & 18.6 & 11862.0 & 1.1 \\
\rowcolor{gray!15} with CrossGET & 78.7 & 62.6 & 51.8 & 71.4 & 58.0 & 84.9 & 1548.8 & 66.3 & 62.0 & 67.5 & 4500.3$_{\color{ForestGreen}\downarrow 47\%}$ & 37.0$_{\color{ForestGreen}\uparrow 99\%}$ & 7825.9$_{\color{ForestGreen}\downarrow 34\%}$ & 1.6$_{\color{ForestGreen}\uparrow 45\%}$ \\
\bottomrule
\end{tabular}
\label{table llava}
\end{table*}

\begin{table*}[t]
    \setlength{\tabcolsep}{0.78mm}
    \captionsetup{font={small}}
    \small
    \centering
    \caption{\textbf{Accelerate CLIP on the CoOp benchmark for the few-shot Image Classification task.} Following the same settings as CoOp, we use 16 shots and report top-1 accuracy on each of the 11 datasets. Details of each dataset are provided in Appendix \ref{section coop benchmark}.}
    \begin{tabular}{lcccccccccccll}
    \toprule
    \rotbox{Approach} & \rotbox{ImageNet} & \rotbox{Caltech101} & \rotbox{OxfordPets} & \rotbox{StanfordCars} & \rotbox{Flowers102} & \rotbox{Food101} & \rotbox{FGVCAircraft} & \rotbox{SUN397} & \rotbox{DTD} & \rotbox{EuroSAT} & \rotbox{UCF101} & \rotbox{\textbf{\textit{Average}}} & \rotbox{GFLOPs} \\
    \cmidrule(r){1-12} \cmidrule{13-14}
    CoOp \cite{zhou2022learning} & \fade{71.1} & \fade{95.4} & \fade{93.3} & \fade{77.5} & \fade{95.6} & \fade{86.5} & \fade{37.3} & \fade{75.1} & \fade{65.8} & \fade{82.6} & \fade{83.7} & 78.5 & 20.6 \\
    \cmidrule(r){1-12} \cmidrule{13-14}
    & \fade{70.8} & \fade{94.6} & \fade{90.8} & \fade{81.9} & \fade{95.8} & \fade{82.0} & \fade{43.7} & \fade{74.1} & \fade{65.7} & \fade{88.4} & \fade{82.2} & 79.1$_{\color{ForestGreen}\uparrow 0.6}$ & 16.5$_{\color{ForestGreen}\downarrow 20\%}$ \\
    & \fade{70.2} & \fade{94.9} & \fade{90.1} & \fade{81.1} & \fade{95.0} & \fade{81.5} & \fade{43.1} & \fade{73.5} & \fade{65.9} & \fade{86.9} & \fade{81.9} & 78.6$_{\color{ForestGreen}\uparrow 0.1}$ & 14.2$_{\color{ForestGreen}\downarrow 31\%}$ \\
    \multirow{-3}{*}{\makecell[l]{CoOp with CrossGET \\ (Ours)}} & \fade{67.6} & \fade{93.9} & \fade{89.5} & \fade{76.6} & \fade{93.3} & \fade{79.7} & \fade{41.3} & \fade{72.1} & \fade{64.2} & \fade{84.5} & \fade{80.5} & 76.7$_{\color{red}\downarrow 1.8}$ & 12.0$_{\color{ForestGreen}\downarrow 42\%}$ \\
    \bottomrule
    \end{tabular}
    \label{table coop}
\end{table*}

\subsection{Experiments on CoOp Benchmark for Few-Shot Image Classification}

We followed the same settings as CoOp \cite{zhou2022learning}, which uses 16 shots and freezes the backbone model CLIP while conducting prompt tuning. The experimental results in Table \ref{table coop} demonstrate that CrossGET achieves notable computational cost savings on few-shot Image Classification task\footnote{We report overall accuracy on all classes as in CoOp, rather than splitting classes into two groups and reporting separate accuracy as in CoCoOp \cite{zhou2022conditional}.}. For example, CrossGET achieves a $31\%$ lossless computational cost saving according to the average top-1 accuracy over 11 datasets. Besides, the performance-cost trade-off on the CoOp benchmark is relatively worse than other experiments we have reported, which should be attributed to 1) most of the model parameters (\ie, the whole backbone) are frozen, resulting in worse convergence than the full-parameter fine-tuning we have used for other experiments; 2) only a portion of the datasets are used for few-shot learning, resulting in more severe overfitting than when using the entire datasets in other experiments.

\section{Conclusion}

In this paper, we introduce \textit{CrossGET}, a general token ensemble framework tailored for accelerating vision-language Transformers. \textit{CrossGET} effectively utilizes bidirectional cross-modal guidance to make informed decisions on token selection and ensemble. Notably, our token-matching method is grounded on an approximate complete-graph matching algorithm, ensuring superior token-matching reliability in comparison to bipartite-graph approaches while maintaining parallelizability for high efficiency. In summary, \textit{CrossGET} provides favorable performance-cost tradeoffs and demonstrates robust applicability, as evidenced through extensive empirical evaluations on a multitude of vision-language tasks, datasets, and model architectures.

\section*{Acknowledgements}

This work was supported by the National Key R$\&$D Program of China (2022YFB4701400/4701402), SSTIC Grant (KJZD20230923115106012), Shenzhen Key Laboratory (ZDSYS20210623092001004), Beijing Key Lab of Networked Multimedia, the National Key R$\&$D Program of China (2022ZD0160201), and Shanghai Artificial Intelligence Laboratory (JF-P23KK00072).
\section*{Impact Statement}

This paper introduces work aimed at advancing the field of model acceleration for vision-language Transformers. It outlines numerous potential positive societal impacts, such as saving electrical energy and reducing carbon dioxide emissions. Concerning negative aspects, while we believe that our work does not explicitly introduce any harmful impacts, it is important to consider potential indirect consequences. These may include reliance on technology that could reduce human involvement in certain tasks or the possibility of misusing accelerated models in ways that were not intended. However, these concerns are not directly linked to the inherent nature of our work.
\bibliography{main}
\bibliographystyle{icml2024}

\newpage
\appendix
\onecolumn
\newpage

\section{Supplementary Experiments and Details}
\label{supp_exps}

\subsection{Diagram of Adding Cross Tokes to Different Models}
\label{independent_framework}

\begin{figure}[htb]
    \centering
    \captionsetup{font={small}}
    \includegraphics[width=1.0\linewidth]{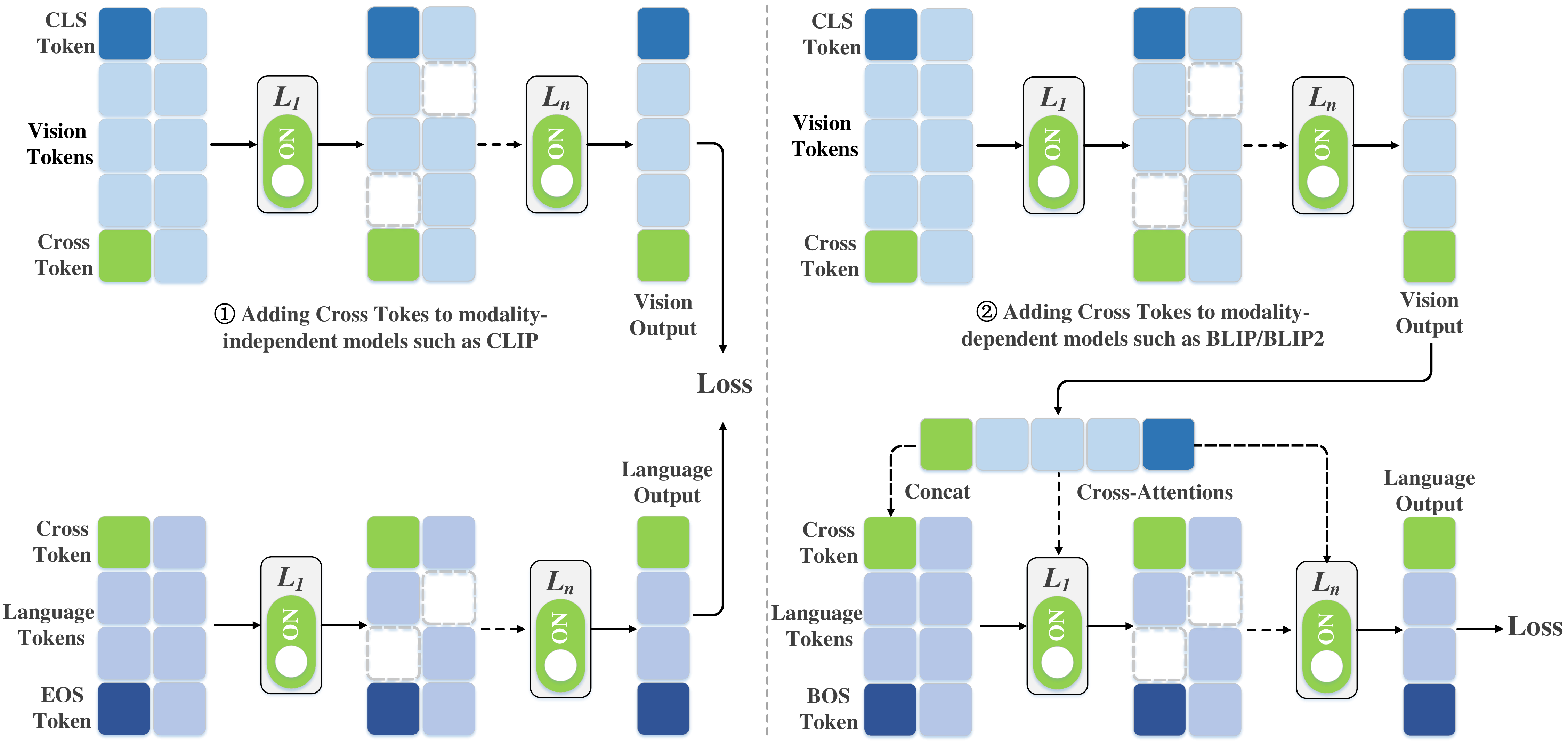}
    \caption{Diagram of adding cross tokes to modality-independent models such as CLIP \cite{radford2021learning} (left) and modality-dependent models such as BLIP/BLIP2 \cite{li2022blip, li2023blip} (right). }
    \label{figure addcross}
\end{figure}

Figure \ref{figure addcross} demonstrates that \textit{CrossGET} is designed to be a general framework that can be used for accelerating both modality-independent vision-language models such as CLIP \cite{radford2021learning} model and modality-dependent vision-language models such as BLIP/BLIP2 \cite{li2022blip, li2023blip} -based models.

Two different types of dependencies exist for modality-dependent vision-language models. The first type that BLIP \cite{li2022blip} belongs to is that the succeeding modality interacts with the final output of the preceding modality through Cross-Attention modules. For this type, in addition to reducing the number of its own tokens, the succeeding modality can also be accelerated by reducing the number of tokens output from the preceding modality to speed up Cross-Attentions.

The second type BLIP2 \cite{li2023blip} -based models such as InstructBLIP \cite{dai2023instructblip}, MiniGPT-4 \cite{zhu2023minigpt}, and mPLUG-Owl \cite{ye2023mplug} belong to is that the succeeding modality takes the final output of the preceding modality as part of its input sequence to the first layer, and the cross-modal interaction is conducted by Self-Attentions in the succeeding modality. For the second type, the succeeding modality can be accelerated by reducing the length of the cross-modal input sequence to speed up Self-Attentions and FFNs in the succeeding modality. 

Take BLIP2 as an example. BLIP2 consists of a ViT for processing visual input, a Q-Former (a Bert) for bridging modalities, and an LLM for taking inputs from Q-Former and generating text output accordingly. During fine-tuning, ViT and Q-Former are tunable while the LLM is frozen. To accelerate BLIP2, we can reduce the number of tokens in both ViT and Q-Former. It is worth noting that there are two ways to reduce the number of tokens processed by LLM. The first one is reducing the number of tokens in Q-Former. Since LLM takes Q-Former’s output as its input, the token reduction conducted in Q-Former leads to LLM acceleration. This setting gives the default performance we reported in the paper. The second one is directly ensembling tokens in LLM. We have also tested this setting and discussed it in Appendix \ref{section apply}.

Recently, vision-language models that take interleaved sequences as inputs, such as Flamingo \cite{alayrac2022flamingo} and VPG-C \cite{li2023fine}, also attracted much attention. CrossGET can be further extended to these models since they can still be categorized as either modality-dependent or modality-independent models. However, some modifications should be made to the concrete strategy for matching tokens. More specifically, for interleaved vision-language inputs, we also need to consider cross-fragment guidance between different image/text fragments within the same modality, which will be more complicated than single image-text scenarios where we only consider cross-modal guidance between different modalities.

\newpage

\subsection{Hyperparameter Settings}

\begin{table*}[!ht]
  \setlength{\tabcolsep}{3 pt}
  \captionsetup{font={small}}
  \small
  \centering
  \caption{Training hyperparameters for accelerating BLIP-based models.}
  \begin{tabular}{l c c c c }
    \toprule
    \multirow{2}{*}[-6pt]{Hyperparameters} & \makecell{BLIP-NLVR \\ \cite{li2022blip}} & \multicolumn{2}{c}{\makecell{BLIP-Captioning \\ \cite{li2022blip}}} & \makecell{BLIP-VQA \\ \cite{li2022blip}} \\ 
    \cmidrule(lr){2-2} \cmidrule(lr){3-4} \cmidrule(lr){5-5}
    & \makecell{NLVR2 \\ \cite{suhr2018corpus}} & \makecell{COCO Caption\\ \cite{chen2015microsoft}} & \makecell{NoCaps \\ \cite{agrawal2019nocaps}} & \makecell{VQAv2 \\ \cite{goyal2017making}} \\
    \midrule
    Optimizer & \multicolumn{4}{c}{AdamW\cite{loshchilov2017decoupled}} \\
    AdamW $\beta$ & \multicolumn{4}{c}{(0.9, 0.999)} \\
    Batch size & \multicolumn{4}{c}{512} \\
    Weight decay & 0.05 & 0.05 & 0.05 & 0.05 \\
    Epochs & 15 & 5 & 5 & 10  \\
    Initial learning rate & $3 \times 10^{-6}$ & $1 \times 10^{-5}$ & $1 \times 10^{-5}$ & $2 \times 10^{-5}$ \\
    Learning rate schedule & \multicolumn{4}{c}{CosineLRScheduler \cite{loshchilov2016sgdr}} \\
    Data augmentation & \multicolumn{4}{c}{RandomAugment \cite{Cubuk_2020_CVPR_Workshops}}\\
    Training Precision & \multicolumn{4}{c}{Mixed Precision \cite{micikevicius2017mixed}}\\
    Matching loss coefficient & 10$^1$ & 10$^2$ & 10$^2$ & 10$^1$ \\
    \bottomrule
  \label{table training hyperparameters blip}
  \end{tabular}
\end{table*}

\begin{table*}[!ht]
  \setlength{\tabcolsep}{4 pt}
  \captionsetup{font={small}}
  \small
  \centering
  \caption{Training hyperparameters for accelerating CLIP and BLIP2-based models.}
  \begin{tabular}{l c c c }
    \toprule
    \multirow{2}{*}[-6pt]{Hyperparameters} & \makecell{CLIP-Retrieval \\ \cite{radford2021learning}} & \makecell{BLIP2-OPT2.7B- \\ Captioning \cite{li2023blip}} & \makecell{BLIP2-OPT6.7B- \\ Captioning \cite{li2023blip}} \\ 
    \cmidrule(lr){2-2} \cmidrule(lr){3-3} \cmidrule(lr){4-4}
    & \makecell{Flickr30K \\ \cite{young2014image}} & \makecell{COCO Caption\\ \cite{chen2015microsoft}} & \makecell{COCO Caption\\ \cite{chen2015microsoft}} \\
    \midrule
    Optimizer & \multicolumn{3}{c}{AdamW \cite{loshchilov2017decoupled}} \\
    AdamW $\beta$ & \multicolumn{3}{c}{(0.9, 0.999)} \\
    Batch size & \multicolumn{1}{c}{512} & 1024 & 512 \\
    Weight decay & 0.2 & 0.05 & 0.05 \\
    Epochs & 12 & 5 & 5 \\
    Initial learning rate & $1 \times 10^{-5}$ & $1 \times 10^{-5}$ & $1 \times 10^{-5}$ \\
    Learning rate schedule & \multicolumn{3}{c}{CosineLRScheduler \cite{loshchilov2016sgdr}} \\
    Data augmentation & \multicolumn{3}{c}{RandomAugment \cite{Cubuk_2020_CVPR_Workshops}} \\
    Training Precision & \multicolumn{3}{c}{Mixed Precision \cite{micikevicius2017mixed}} \\
    Matching loss coefficient & 10$^0$ & 10$^{-1}$ & 10$^{-1}$ \\
    \bottomrule
  \label{table training hyperparameters clip and blip2}
  \end{tabular}
\end{table*}

\begin{table*}[!ht]
  \setlength{\tabcolsep}{4 pt}
  \captionsetup{font={small}}
  \small
  \centering
  \caption{Training hyperparameters for accelerating CoOp and LLaVA-1.5 models.}
  \begin{tabular}{l c c c }
    \toprule
    \multirow{2}{*}[-6pt]{Hyperparameters} & \makecell{CLIP-CoOp \\ \cite{zhou2022learning}} & \makecell{LLaVA-1.5-7B\\ \cite{liu2023improved}} & \makecell{LLaVA-1.5-13B\\\cite{liu2023improved}} \\ 
    \cmidrule(lr){2-2} \cmidrule(lr){3-3} \cmidrule(lr){4-4}
    & \makecell{See Appendix \ref{section coop benchmark}} & \makecell{SFT data of LLaVA-1.5} & \makecell{SFT data of LLaVA-1.5} \\
    \midrule
    Optimizer & \multicolumn{3}{c}{AdamW \cite{loshchilov2017decoupled}} \\
    AdamW $\beta$ & \multicolumn{3}{c}{(0.9, 0.999)} \\
    Batch size & \multicolumn{1}{c}{EuroSAT: 128. Others: 256} & 128 & 128 \\
    Weight decay & 0.0005 & 0 & 0 \\
    Epochs & ImageNet: 50. Others: 200 & 1 & 1 \\
    Initial learning rate & ImageNet: $2 \times 10^{-2}$. Others: $5 \times 10^{-2}$ & $2 \times 10^{-5}$ & $2 \times 10^{-5}$ \\
    Learning rate schedule & \multicolumn{3}{c}{CosineLRScheduler \cite{loshchilov2016sgdr}} \\
    Matching loss coefficient & 10$^1$ & 10$^{-3}$ & 10$^{-1}$ \\
    \bottomrule
  \label{table training hyperparameters coop and llava}
  \end{tabular}
\end{table*}

\begin{table*}[!ht]
  \setlength{\tabcolsep}{2.7 pt}
  \captionsetup{font={small}}
  \small
  \centering
  \caption{Structure hyperparameters for all models used in our experiments. The superscript $\textsuperscript{*}$ indicates 2 Transformers share parameters. The superscript \textsuperscript{$\dagger$} indicates hyperparameters are from (OPT, Q-Former).}
  \begin{tabular}{lccccccccc}
    \toprule
    \multirow{2}{*}{Model} & \multirow{2}{*}{\makecell{Input\\resolution}} & \multicolumn{4}{c}{\makecell{Vision Transformer \\ \cite{touvron2021training, fang2023eva}}} & \multicolumn{4}{c}{\makecell{Language Transformer \\ \cite{devlin2018bert, zhang2022opt}}} \\
    \cmidrule(lr){3-6} \cmidrule{7-10}
    & & number & layers & width & heads & number & layers & width & heads \\
    \midrule
    CLIP-Retrieval & 336$\times$336 & 1 & 12 & 768 & 12 & 1 & 12 & 512 & 8 \\
    CLIP-CoOp & 336$\times$336 & 1 & 12 & 768 & 12 & 1 & 12 & 512 & 8 \\
    BLIP-NLVR & 384$\times$384 & 2$\textsuperscript{*}$ & (12, 12) & (768, 768) & (12, 12) & 1 & 12 & 768 & 12 \\
    BLIP-Captioning & 384$\times$384 & 1 & 12 & 768 & 12 & 1 & 12 & 768 & 12 \\
    BLIP-NoCaps & 384$\times$384 & 1 & 12 & 768 & 12 & 1 & 12 & 768 & 12 \\
    BLIP-VQA & 480$\times$480 & 1 & 12 & 768 & 12 & 2 & (12, 12) & (768, 768) & (12, 12) \\
    BLIP2-OPT2.7B & 364$\times$364 & 1 & 39 & 1408 & 16 & 2\textsuperscript{$\dagger$} & (32, 12) & (2560, 768) & (32, 12) \\
    BLIP2-OPT6.7B & 364$\times$364 & 1 & 39 & 1408 & 16 & 2\textsuperscript{$\dagger$} & (32, 12) & (4096, 768)& (32, 12) \\
    LLaVA-1.5-7B & 336$\times$336 & 1 & 24 & 1024 & 16 & 1 & 32 & 4096 & 32 \\
    LLaVA-1.5-13B & 336$\times$336 & 1 & 24 & 1024 & 16 & 1 & 40 & 5120 & 40 \\
    \bottomrule
  \label{table structure hyperparameters}
  \end{tabular}
\end{table*}

The hyperparameters about model training are listed in Table \ref{table training hyperparameters blip}, Table \ref{table training hyperparameters clip and blip2}, and Table \ref{table training hyperparameters coop and llava}. The hyperparameters about model structures are listed in Table \ref{table structure hyperparameters}.

\subsection{Evaluation Datasets for CoOp}
\label{section coop benchmark}

The CoOp benchmark \cite{zhou2022learning} consists of 11 datasets, which are ImageNet (1000 classes) \cite{deng2009imagenet}, Caltech101 (100 classes) \cite{fei2004learning}, OxfordPets (37 classes) \cite{parkhi2012cats}, StanfordCars (196 classes) \cite{krause20133d}, Flowers102 (102 classes) \cite{nilsback2008automated}, Food101 (101 classes) \cite{bossard2014food}, FGVCAircraft (100 classes) \cite{maji2013fine}, SUN397 (397 classes) \cite{xiao2010sun}, DTD (47 classes) \cite{cimpoi2014describing}, EuroSAT (10 classes) \cite{helber2019eurosat}, and UCF101 (101 classes) \cite{soomro2012ucf101}. For each dataset, we randomly sample 16 images in each class as its training set for few-shot learning.

\subsection{Evaluation Datasets for LLaVA-1.5}
\label{section llava benchmark}

After applying CrossGET to LLaVA-1.5 \cite{liu2023improved}, we used 10 datasets to evaluate model performance, including VQA-v2 \cite{goyal2017making}, GQA \cite{hudson2019gqa}, VisWiz \cite{gurari2018vizwiz}, ScienceQA-IMG \cite{lu2022learn}, TextVQA \cite{singh2019towards}, POPE \cite{li2023evaluating}, MME \cite{yin2023survey}, MMBench \cite{liu2023mmbench}, MMBench-CN \cite{liu2023mmbench}, and SEED-Bench-Image \cite{li2023seed}. 

\subsection{Ablation Study on Training Hyperparameters}
\label{section training hyperparameters}

The hyperparameters are basically inherited from original models and do not need a specific tune. The particular case is that batch sizes are adjusted to fit our computational resources. The only additional hyperparameter introduced by \textit{CrossGET} is the matching loss coefficient $\alpha$, which is used to balance the original loss items and the matching loss item. For simplicity, $\alpha$ can be determined as the number $\alpha \in \{10^{i}\}_{i\in \mathbb{N}}$ that makes the original loss items and the matching loss item have the closest order of magnitude, and therefore, it does not need to be tuned either.

\begin{table*}[htb]
\captionsetup{font={small}}
\small
\setlength{\tabcolsep}{1.4mm}
    \begin{minipage}{0.31\linewidth}
    \caption{Ablation study about batch size on BLIP-NLVR.}
    \vspace{-6pt}
        \flushleft
        \begin{tabular}{l @{\hspace{1.0\tabcolsep}} l @{\hspace{1.0\tabcolsep}} l }
        \toprule
        Batch size & Dev Acc & Test Acc \\
        \cmidrule(r){1-1} \cmidrule{2-3}
         $128$ & \fade{82.0}$_{\color{red}\downarrow 0.1}$ & 82.8$_{\color{red}\downarrow 0.4}$ \\
         $256$ & \textbf{\fade{82.2}}$_{\color{ForestGreen}\uparrow 0.1}$ & 83.0$_{\color{red}\downarrow 0.2}$  \\
         \rowcolor{gray!10} $512$ & \fade{82.1} & \textbf{83.2}  \\
         $1024$ & \textbf{\fade{82.2}}$_{\color{ForestGreen}\uparrow 0.1}$ & 83.0$_{\color{red}\downarrow 0.2}$  \\
        \bottomrule
        \end{tabular}
      \label{table batch size}
    \end{minipage}
    \hspace{0.01\linewidth}
    \medskip
    \begin{minipage}{0.31\linewidth}
    \caption{Ablation study about learning rate on BLIP-NLVR.}
    \vspace{-6pt}
        \flushleft
        \begin{tabular}{l @{\hspace{1.0\tabcolsep}} l @{\hspace{1.0\tabcolsep}} l }
        \toprule
        Learning rate & Dev Acc & Test Acc \\
        \cmidrule(r){1-1} \cmidrule{2-3}
        $1\times 10^{-6}$ & \fade{81.8}$_{\color{red}\downarrow 0.3}$ & 82.5$_{\color{red}\downarrow 0.7}$  \\
        \rowcolor{gray!10}  $3\times 10^{-6}$ & \fade{82.1} & \textbf{83.2} \\
        $1\times 10^{-5}$ & \textbf{\fade{82.2}}$_{\color{ForestGreen}\uparrow 0.1}$ & 82.7$_{\color{red}\downarrow 0.5}$ \\
        $3\times 10^{-5}$ & \fade{82.0}$_{\color{red}\downarrow 0.1}$ & 82.6$_{\color{red}\downarrow 0.6}$  \\
        \bottomrule
        \end{tabular}
      \label{table learning rate}
    \end{minipage}
    \hspace{0.04\linewidth}
    \medskip
    \begin{minipage}{0.29\linewidth}
    \caption{Ablation study about coefficient $\alpha$ for matching loss on BLIP-NLVR.}
    \vspace{-6pt}
        \flushleft
        \begin{tabular}{l @{\hspace{1.0\tabcolsep}} l @{\hspace{1.0\tabcolsep}} l }
        \toprule
        Coefficient & Dev Acc & Test Acc \\
        \cmidrule(r){1-1} \cmidrule{2-3}
         $10^{0}$ & \fade{82.0}$_{\color{red}\downarrow 0.1}$ & 82.5$_{\color{red}\downarrow 0.7}$ \\
         \rowcolor{gray!10} $10^{1}$ & \textbf{\fade{82.1}} & \textbf{83.2} \\
         $10^{2}$ & \fade{81.8}$_{\color{red}\downarrow 0.3}$ & 82.7$_{\color{red}\downarrow 0.5}$  \\
        \bottomrule
        \end{tabular}
      \label{table coefficient}
    \end{minipage} 
\end{table*}

Table \ref{table batch size}, Table \ref{table learning rate}, and Table \ref{table coefficient} investigate how hyperparameters affect the model performance. Experimental results show that the performance is insensitive to batch size and slightly sensitive to the learning rate. As for the matching loss coefficient $\alpha \in \{10^{i}\}_{i\in \mathbb{N}}$, set it to the value that makes the original loss items and the matching loss item have the closest order of magnitude as mentioned above will work well.

\subsection{Ablation Study on Different Modalities}
\label{section ablation modalities}

\begin{table*}[h]
\captionsetup{font={small}}
\small
\centering
\caption{Ablation study about applying \textit{CrossGET} on different modalities.}
\vspace{-6pt}
    \begin{tabular}{l l l l}
    \toprule
    Modality  & I2T R@1 & T2I R@1 & GFLOPs  \\
    \cmidrule(r){1-1} \cmidrule(r){2-3} \cmidrule{4-4}
     \rowcolor{gray!10} vision only & 92.1 & 79.6 & 12.0  \\
     language only & \textbf{92.8}$_{\color{ForestGreen}\uparrow 0.7}$ & \textbf{80.4}$_{\color{ForestGreen}\uparrow 0.8}$ & 19.3$_{\color{red}\uparrow 61\%}$  \\
     vision and language & 91.4$_{\color{red}\downarrow 0.7}$ & 78.3$_{\color{red}\downarrow 1.3}$ & \textbf{10.6}$_{\color{ForestGreen}\downarrow 12\%}$  \\
    \bottomrule
    \end{tabular}
\label{table ablation modality}
\end{table*}

As shown in Figure \ref{figure cross}, it is flexible that \textit{CrossGET} can be applied on both vision and language modalities or only on one of the modalities. Table \ref{table ablation modality} investigates the trade-off between model performance and computational cost of application on different modalities. Experimental results show that \textit{CrossGET} only on the vision modality achieves the best trade-off.

\subsection{Ablation Study on the Strategy of Adding Cross Token}

\begin{table*}[h]
\captionsetup{font={small}}
\small
\centering
\caption{Ablation study about the strategy of adding cross tokens. }
\vspace{-6pt}
    \begin{tabular}{l l l l }
    \toprule
    Depth  & I2T R@1 & T2I R@1 & GFLOPs  \\
    \cmidrule(r){1-1} \cmidrule(r){2-3} \cmidrule{4-4}
     shallow & 91.5$_{\color{red}\downarrow 0.6}$ & 79.5$_{\color{red}\downarrow 0.1}$ & 12.0  \\
     \rowcolor{gray!10} deep & \textbf{92.1} & \textbf{79.6} & 12.0  \\
     share & 90.7$_{\color{red}\downarrow 1.4}$ & 78.8$_{\color{red}\downarrow 0.8}$ & 12.0  \\
    \bottomrule
    \end{tabular}
\label{table ablation depth}
\end{table*}

There are several strategies for injecting cross tokens into the model. For example, (1) deep: adding different cross tokens for each layer; (2) shallow: only adding one cross token into the first layer; (3) share: adding one cross token but jointly optimized in each layer. Table \ref{table ablation depth} shows that adding different cross tokens for each layer achieves the best performance. 

\subsection{Ablation Study on the Initialization of Cross Token}
\label{section initialization}

\begin{table*}[h]
\captionsetup{font={small}}
\small
\centering
\caption{Ablation study about initializing cross tokens.}
\vspace{-6pt}
    \begin{tabular}{l l l l}
    \toprule
    Initialization  & I2T R@1 & T2I R@1 & GFLOPs  \\
    \cmidrule(r){1-1} \cmidrule(r){2-3} \cmidrule{4-4}
     zero & 91.7$_{\color{red}\downarrow 0.4}$ & 77.9$_{\color{red}\downarrow 1.7}$ & 12.0 \\
     normal random & 90.4$_{\color{red}\downarrow 1.7}$ & 77.6$_{\color{red}\downarrow 2.0}$ & 12.0  \\
     uniform random & 90.2$_{\color{red}\downarrow 1.9}$ & 77.6$_{\color{red}\downarrow 2.0}$ & 12.0  \\
     \rowcolor{gray!10} $\text{informative tokens}$ & \textbf{92.1} & \textbf{79.6} & 12.0  \\
    \bottomrule
    \end{tabular}
\label{table ablation initialization}
\end{table*}

For fine-tuning, cross tokens are kind of sensitive to the initialization strategy. Using informative tokens to initialize cross tokens is recommended. More specifically, for the vision modality, [CLS] token can be used to initialize the cross token. For the language modality, the cross token can be initialized by [CLS]/[EOS]/[EOT] tokens for discriminative tasks (it depends on which token is ultimately used to calculate the loss) and by [BOS] token for auto-regressive tasks (if there is no, we use the first token of the input sequence to initialize instead).

Table \ref{table ablation initialization} shows that zero initialization and random initialization perform worse. We think the sensitivity should be attributed to the limited training time for fine-tuning and the purpose of quickly adapting to downstream tasks. More specifically, random/zero initialization may work well for pre-training since there is enough time for cross tokens to learn informative guidance. However, it will be difficult for random/zero initialization to learn well with limited iterations for fine-tuning. Therefore, initializing the cross token with [CLS] token in the vision modality and [CLS]/[BOS]/[EOS]/[EOT] token in the language modality implies that the cross token already contains some informative guidance of the modality it is in, and would be easier to form more informative guidance with this good starting point.

\subsection{Ablation Study on the Projection Layer Detach}

\begin{table*}[h]
\captionsetup{font={small}}
\small
\centering
\caption{Ablation study about projection layer detach.}
\vspace{-6pt}
    \begin{tabular}{l l l l }
    \toprule
    Projection detach  & I2T R@1 & T2I R@1 & GFLOPs  \\
    \cmidrule(r){1-1} \cmidrule(r){2-3} \cmidrule{4-4}
     neither & 91.5$_{\color{red}\downarrow 0.6}$ & 78.9$_{\color{red}\downarrow 0.7}$ & 12.0  \\
     vision only & 91.4$_{\color{red}\downarrow 0.7}$ & 79.4$_{\color{red}\downarrow 0.2}$ & 12.0  \\
     language only & 90.5$_{\color{red}\downarrow 1.6}$ & 78.9$_{\color{red}\downarrow 0.7}$ & 12.0  \\
     \rowcolor{gray!10} both & \textbf{92.1} & \textbf{79.6} & 12.0 \\
    \bottomrule
    \end{tabular}
\label{table projection detach}
\end{table*}

The final projection layers are initially used to project features from the different modalities into aligned representations. In \textit{CrossGET}, the final projection layers are detached from the original model and used for aligning cross tokens. The detach operation prevents gradients with respect to cross tokens from updating the projection layers. Table \ref{table projection detach} shows that detaching both vision and language projection improves performance.

\subsection{Ablation Study on the Number of Cross Tokens}
\label{section ablation number of cross tokens}

\begin{table*}[h]
\captionsetup{font={small}}
\small
\caption{Ablation study on number of cross tokens.}
    \vspace{-6pt}
    \centering
    \begin{tabular}{l l l l }
    \toprule
    Number & Dev Acc & Test Acc & GFLOPs \\
    \cmidrule(r){1-1} \cmidrule(r){2-3} \cmidrule{4-4}
    \rowcolor{gray!10} $1$  & \fade{82.1} & \textbf{83.2} & \textbf{61.1} \\
    $2$ & \textbf{\fade{82.2}}$_{\color{ForestGreen}\uparrow 0.1}$ & \textbf{83.2}$_{\color{ForestGreen}\uparrow 0.0}$ & 61.4$_{\color{red}\uparrow 0.3}$ \\
    $3$ & \fade{81.9}$_{\color{red}\downarrow 0.2}$ & \textbf{83.2}$_{\color{ForestGreen}\downarrow 0.0}$ & 61.8$_{\color{red}\uparrow 0.7}$\\
    $4$ & \fade{82.0}$_{\color{red}\downarrow 0.1}$ & 82.9$_{\color{red}\downarrow 0.3}$ & 62.2$_{\color{red}\uparrow 1.1}$ \\
    \bottomrule
    \end{tabular}
\label{table cross tokens}
\end{table*}

Table \ref{table cross tokens} investigates how the performance is impacted by the number of cross tokens on the Vision Reasoning Task and BLIP \cite{li2022blip} model. It can be observed that the performance is not sensitive to the increase in the number of tokens, which is unlike prompt tuning \cite{lester2021power, jia2022visual} that model performance can be boosted by increasing the number of tokens. Considering the additional computational cost of multiple cross tokens, using only one cross token is recommended.

\subsection{Ablation Study on Tokens for Computing Importance}
\label{section ablation computing importance}

For BLIP \cite{li2022blip} on the Visual Reasoning task, a different setting from default is that not cross tokens alone, but all tokens are used to compute importance. By default, in the modality-independent model CLIP \cite{radford2021learning}, only the [CLS] and [EOS] tokens are ultimately used for computing loss. In contrast, for the modality-dependent model BLIP-NLVR, all tokens output from the vision modality are parts of the inputs for the language modality and matter. 

\begin{table*}[h]
\captionsetup{font={small}}
\small
\setlength{\tabcolsep}{2mm}
\centering
\caption{Ablation study about tokens for computing importance.}
    \begin{tabular}{l l l }
    \toprule
    used tokens & Dev Acc & Test Acc   \\
    \cmidrule(r){1-1} \cmidrule{2-3}
    cross token & \fade{82.1}$_{\color{ForestGreen}\uparrow 0.0}$ & 82.9$_{\color{red}\downarrow 0.3}$  \\
    other tokens & \fade{81.9}$_{\color{red}\downarrow 0.2}$ & 82.2$_{\color{red}\downarrow 1.0}$  \\
    \rowcolor{gray!10} all tokens & \fade{82.1} & \textbf{83.2} \\
    importance & \textbf{\fade{82.2}}$_{\color{ForestGreen}\uparrow 0.1}$ & 83.0$_{\color{red}\downarrow 0.2}$  \\
    \bottomrule
\end{tabular}
\label{table nlvr ablation}
\end{table*}

Four settings about which tokens are used for computing importance are tested as shown in Table \ref{table nlvr ablation}: (1) cross token: cross tokens contribute all; (2) other tokens: other tokens contribute all; (3) all tokens (adopted): cross tokens contribute to $\frac{1}{2}$ importance while other tokens contribute to the other $\frac{1}{2}$; (4) importance: we can reuse the dot product between the query and key of each token (including cross tokens) that has already been calculated in the Self-Attention as the importance metric to avoid extra computational cost for introducing other tokens' importance. 

\subsection{Ablation Study on Tokens for Computing JS divergence}
\label{section ablation computing js divergence}

\begin{table*}[h]
\captionsetup{font={small}}
\small
\setlength{\tabcolsep}{2mm}
\centering
\caption{Ablation study about which tokens are used for computing JS divergence as additional loss items.}
    \begin{tabular}{l l l}
    \toprule
    JS divergence as loss  & CIDEr & SPICE  \\
    \cmidrule(r){1-1} \cmidrule{2-3}
    only between pairs of cross tokens & 130.2$_{\color{red}\downarrow 1.4}$ & 23.7$_{\color{red}\downarrow 0.1}$\\
    only between cross tokens and other tokens & 131.0$_{\color{red}\downarrow 0.6}$ & 23.5$_{\color{red}\downarrow 0.3}$ \\
    w/o weighting loss according to generation order & 131.2$_{\color{red}\downarrow 0.4}$ & 23.7$_{\color{red}\downarrow 0.1}$ \\
    \rowcolor{gray!10} between cross tokens and all tokens & \textbf{131.6} & \textbf{23.8} \\
    \bottomrule
    \end{tabular}
    \label{table caption coco ablation}
\end{table*}

For BLIP \cite{li2022blip} on the Image Caption task, a different setting from default is that not only the loss of JS divergence between the pairs of cross tokens but also the JS divergence between the cross tokens and other tokens should be added as loss items. By default, in the language modality of the discriminative model CLIP \cite{radford2021learning}, only the [EOS] token matters for the final output. In contrast, for the auto-regressive model BLIP-Captioning, tokens are generated based on their previous tokens, and therefore, every token matters. 

Table \ref{table caption coco ablation} shows that combined JS divergence between pairs of cross tokens as well as between cross tokens and other tokens as loss performs best. Besides, weighting the loss between cross tokens and other tokens according to the generation order also helps. The weight for the $i$-th generated token is $1 - \frac{i}{L}$ where $L$ is the maximum generation length, which means the first generated token is more important than the later ones since they are generated based on former ones. 

\subsection{Comparison Experiments with Text-Relevant Image Patch Selection}
\label{appendix trips}

\begin{table*}[h]
  \setlength{\tabcolsep}{0.25mm}
  \captionsetup{font={small}}
  \small
  \centering
  \caption{\textbf{Accelerate CLIP on the Flickr30K dataset of the Image-Text Retrieval task.} R: Recall. R@1, R@5, and R@10 are the higher the better. The TRIPS represents Text-Relevant Image Patch Selection \cite{jiang2022trips}. The -L indicates using an additional learnable projection to align the text [CLS] token with vision tokens.}
  \begin{tabular}{ l @{\hspace{2\tabcolsep}} l l l l l l l l l}
    \toprule
    \multirow{2}{*}[-3pt]{Approach} & \multicolumn{3}{c}{Image $\rightarrow$ Text} & \multicolumn{3}{c}{Text $\rightarrow$ Image} & \multicolumn{1}{c}{Avg.} &\multirow{2}{*}[-3pt]{\makecell{GFLOPs \\ $\downarrow$}} & \multirow{2}{*}[-3pt]{\makecell{Throughput \\ $\uparrow$}} \\ 
    \cmidrule(r){2-4} \cmidrule(lr){5-7} \cmidrule(r){8-8}
    & R@1 & R@5 & R@10 & R@1 & R@5 & R@10 & $\overline{\textbf{R@1}}$ \\
    \cmidrule{1-10}
    CLIP \cite{radford2021learning} & 92.1 & \fade{99.1} & \fade{99.7} & 79.3 & \fade{95.7} & \fade{98.0} & 85.7 & 20.6 & 255.2 \\
    \cmidrule{1-10}
    TRIPS (Default FLOPs) & 87.6 & \fade{98.7} & \fade{99.4} & 76.6 & \fade{94.4} & \fade{97.0} & 82.1 & 16.4 & 317.7 \\
    TRIPS-L (Default FLOPs) & 90.4 & \fade{98.9} & \fade{99.5} & 76.8 & \fade{94.4} & \fade{97.2} & 83.6 & 16.4 & 316.9 \\
    TRIPS (Same FLOPs) & 75.5 & \fade{94.3} & \fade{97.8} & 63.9 & \fade{88.5} & \fade{93.8} & 69.7 & 12.0 & 423.5  \\
    TRIPS-L (Same FLOPs) & 70.1 & \fade{92.4} & \fade{96.9} & 61.2 & \fade{86.8} & \fade{92.1} & 65.7 & 12.0 & 423.1 \\
    \cmidrule{1-10}
    \rowcolor{gray!10}CrossGET (Ours) & \textbf{92.1}$_{\color{ForestGreen}\uparrow 0.0}$ & \textbf{\fade{99.7}}$_{\color{ForestGreen}\uparrow 0.6}$ & \textbf{\fade{99.8}}$_{\color{ForestGreen}\uparrow 0.1}$ & \textbf{79.6}$_{\color{ForestGreen}\uparrow 0.3}$ & \textbf{\fade{95.7}}$_{\color{ForestGreen}\uparrow 0.0}$ & \textbf{\fade{98.0}}$_{\color{ForestGreen}\uparrow 0.0}$ & \textbf{85.9}$_{\color{ForestGreen}\uparrow 0.2}$ & 12.0$_{\color{ForestGreen}\downarrow 42\%}$ & 401.8$_{\color{ForestGreen}\uparrow 57\%}$\\
    \bottomrule
  \end{tabular}
  \label{table comparison trips}
\end{table*}

In Table \ref{table comparison trips}, the TRIPS (Default FLOPs) \cite{jiang2022trips} indicates we follow the recommended setting of the original TRIPS, \ie, we take the 5th and 10th as the patch-selection layer and set the keep ratio of each layer to 70$\%$. The TRIPS (Same FLOPs) indicates we decrease the keep ratio of each patch-selection layer to achieve similar GFLOPs with ToMe \cite{bolya2022tome} and CrossGET. Overall, the experimental results demonstrate that CrossGET outperforms TRIPS under similar computational costs.

When compared with TRIPS, one of the \textit{CrossGET}'s advantages is that it can more easily deal with the models in which the embedding sizes of vision and language branches are different. More specifically, TRIPS is not directly applicable to the models with different embedding sizes of the vision branch and language branch, and without a projection layer that projects the language embedding size into the vision embedding size as well.

For example, the vision and language embedding sizes in the CLIP model we used are 768 and 512, respectively. Besides, there is a 768 $\rightarrow$ 512 projection layer for vision projection and a 512 $\rightarrow$ 512 for language projection. TRIPS requires the projected text [CLS] token to have the same embedding size as the tokens in the vision branch. However, CLIP has no trained (\ie, aligned) 512 $\rightarrow$ 768 projection layer to fulfill this. To overcome this problem, we propose two strategies: 1) The first one is to use the pseudo-inverse of the trained 768 $\rightarrow$ 512 projection layer to project the 512-dimensional text [CLS] token into a 768-dimensional token, whose experimental results are denoted without -L. 2) The second one is to add an additional 512 $\rightarrow$ 768 learnable projection layer into the original model and then jointly optimize, whose experimental results are denoted with -L. Note that this is not a problem for \textit{CrossGET} since cross tokens are learned cross-modally while used intra-modally, and the embedding size of cross tokens is the same as other tokens in the same modality branch. Thus, CrossGET doesn’t need a projection layer to align cross tokens when they are used as metrics to guide the token reduction.

The other advantage of CrossGET is that it can deal with both modality-independent and modality-dependent models, which is also an important contribution of CrossGET. We have discussed this in Section \ref{Cross-guided matching}, and TRIPS can serve as an example to elaborate it. More specifically, TRIPS uses text [CLS] token, \ie, the output of the language encoder in the ALBEF \cite{li2021align} model as the metric to guide the token reduction in the vision branch. However, this paradigm cannot be used in multimodal models where the input of the language branches depends on the output of the vision branch.

For example, in the BLIP-NLVR \cite{li2022blip} model, the output of the vision branch is a necessary input for the language branch. And if we want to use the text [CLS] token \ie, the output of the language branch as a metric to guide the token reduction in the vision branch, we have to first forward through the vision branch, get the last layer’s output as the input of the language branch, forward through the language branch, get the last layer’s output as the metric, \ie, only after the forward of the vision branch is finished, we can get the required metric used for vision branch. CrossGET breaks this paradox of cycles by using cross tokens as proxies for other modalities, providing cross-modal guidance on behalf of other modalities without being constrained by the order of calculations.

\subsection{Comparison Experiments with Adpater}
\label{section adapter}

\begin{table*}[h]
  \setlength{\tabcolsep}{0.25mm}
  \captionsetup{font={small}}
  \small
  \centering
  \caption{\textbf{Accelerate CLIP on the Flickr30K dataset of the Image-Text Retrieval task.} R: Recall. R@1, R@5, and R@10 are the higher the better. The Adapter-x represents Adaptformer \cite{chen2022adaptformer}, and the integer x in Adapter-x represents the middle dimension of the adapter. }
  \begin{tabular}{ l @{\hspace{2\tabcolsep}} l l l l l l l l l}
    \toprule
    \multirow{2}{*}[-3pt]{Approach} & \multicolumn{3}{c}{Image $\rightarrow$ Text} & \multicolumn{3}{c}{Text $\rightarrow$ Image} & \multicolumn{1}{c}{Avg.} &\multirow{2}{*}[-3pt]{\makecell{GFLOPs \\ $\downarrow$}} & \multirow{2}{*}[-3pt]{\makecell{Throughput \\ $\uparrow$}} \\ 
    \cmidrule(r){2-4} \cmidrule(lr){5-7} \cmidrule(r){8-8}
    & R@1 & R@5 & R@10 & R@1 & R@5 & R@10 & $\overline{\textbf{R@1}}$ \\
    \cmidrule{1-10}
    CLIP \cite{radford2021learning} & 92.1 & \fade{99.1} & \fade{99.7} & 79.3 & \fade{95.7} & \fade{98.0} & 85.7 & 20.6 & 255.2 \\
    \cmidrule{1-10}
    ToMe \cite{bolya2022tome}  & 90.8 & \fade{99.2} & \fade{99.5} & 78.1 & \fade{95.3} & \fade{97.7} & 84.5 & 11.8  & 417.4 \\
    ToMe$^{\snowflake}$+Adapter-16$^{\fire}$ & 89.2 & \fade{98.7} & \fade{99.6} & 75.9 & \fade{94.2} & \fade{97.1} & 82.6 & 11.9 & 404.1  \\
    ToMe$^{\snowflake}$+Adapter-64$^{\fire}$ & 89.9 & \fade{98.8} & \fade{99.5} & 76.7 & \fade{94.4} & \fade{97.3} & 83.3 & 12.0 & 401.2 \\
    ToMe$^{\snowflake}$+Adapter-256$^{\fire}$ & 90.2 & \fade{99.0} & \fade{99.4} & 76.7 & \fade{94.5} & \fade{97.5} & 83.5 & 12.5 & 386.4 \\
    ToMe$^{\snowflake}$+Adapter-1024$^{\fire}$ & 90.3 & \fade{99.0} & \textbf{\fade{99.8}} & 78.0 & \fade{94.6} & \fade{97.4} & 84.2 & 14.5 & 346.2 \\
    ToMe$^{\snowflake}$+Adapter-4096$^{\fire}$ & 91.4 & \fade{98.8} & \fade{99.6} & 78.2 & \fade{95.0} & \fade{97.6} & 84.8 & 22.7 & 243.5 \\
    \cmidrule{1-10}
    \rowcolor{gray!10}CrossGET (Ours) & \textbf{92.1}$_{\color{ForestGreen}\uparrow 0.0}$ & \textbf{\fade{99.7}}$_{\color{ForestGreen}\uparrow 0.6}$ & \textbf{\fade{99.8}}$_{\color{ForestGreen}\uparrow 0.1}$ & \textbf{79.6}$_{\color{ForestGreen}\uparrow 0.3}$ & \textbf{\fade{95.7}}$_{\color{ForestGreen}\uparrow 0.0}$ & \textbf{\fade{98.0}}$_{\color{ForestGreen}\uparrow 0.0}$ & \textbf{85.9}$_{\color{ForestGreen}\uparrow 0.2}$ & 12.0$_{\color{ForestGreen}\downarrow 42\%}$ & 401.8$_{\color{ForestGreen}\uparrow 57\%}$\\
    \bottomrule
  \end{tabular}
  \label{table comparison adapter}
\end{table*}

The experimental results demonstrate that when using ToMe \cite{bolya2022tome} with an adapter \cite{chen2022adaptformer}, the middle dimension of the adapter needs to be very large(\eg, around 4096) for the model to perform better than without using the adapter. However, the additional computation cost introduced by the adapter is significant (see GFLOPs and Throughput in the above Table \ref{table comparison adapter}), and the performance is still worse than CrossGET.

\subsection{Experiments with BLIP2 on Image Captioning}
\label{section opt}

\begin{table*}[h]
 \setlength{\tabcolsep}{1.2mm}
  \captionsetup{font={small}}
  \small
  \centering
  \caption{Accelerate multimodal LLM BLIP2-OPT2.7B \cite{li2023blip} on the COCO Caption dataset of the Image Caption task. The suffix -F denotes GFLOPs and throughput for the forward, while -G denotes GFLOPs and throughput for the generation. \textsuperscript{$*$} indicates using greedy decoding instead of beam search for generation. }
  \begin{tabular}{l l l l l l l l l}
    \toprule
    Approach & Tuning & CIDEr & BLEU@4 & GFLOPs-F & Throughput.-F & GFLOPs-G  & Throughput.-G & Throughput.-G\textsuperscript{$*$} \\
    \cmidrule(r){1-2} \cmidrule(r){3-4} \cmidrule(r){5-6} \cmidrule{7-9}
    BLIP2-OPT2.7B & - & 145.6 & 42.8 & 854.2 & 54.0 & 1379.3 & 22.3 & 52.4 \\
    \cmidrule(r){1-2} \cmidrule(r){3-4} \cmidrule(r){5-6} \cmidrule{7-9}
    & \fade{w/o tuning} & \fade{145.1} & \fade{42.6} & \fade{769.1} & \fade{-} & \fade{1294.2} & \fade{-} & \fade{-} \\
    & \fade{w/o tuning} & \fade{144.2} & \fade{42.3} & \fade{679.7} & \fade{-} & \fade{1218.1} & \fade{-} & \fade{-} \\
    & \fade{w/o tuning} & \fade{142.8} & \fade{42.2} & \fade{592.2} & \fade{-} & \fade{1104.0} & \fade{-} & \fade{-} \\
    & \fade{w/o tuning} & \fade{136.5} & \fade{40.6} & \fade{506.7} & \fade{-} & \fade{1018.5} & \fade{-} & \fade{-} \\
    \cmidrule(r){2-2} \cmidrule(r){3-4} \cmidrule(r){5-6} \cmidrule{7-9}
    \multirow{-5}{*}{\makecell[l]{ToMe \\ \cite{bolya2022tome}}} & w/ tuning & 142.4$_{\color{red}\downarrow 3.2}$ & 41.7$_{\color{red}\downarrow 1.1}$ & 404.6 & 107.5 & 855.1 & 30.5 & 86.7 \\
    \midrule
    & \fade{w/o tuning} & \fade{145.9} & \textbf{\fade{43.1}} & \fade{785.4} & \fade{-} & \fade{1310.5} & \fade{-} & \fade{-} \\
    & \fade{w/o tuning} & \fade{144.6} & \textbf{\fade{42.6}} & \fade{692.7} & \fade{-} & \fade{1204.5} & \fade{-} & \fade{-} \\
    & \fade{w/o tuning} & \fade{144.2} & \textbf{\fade{42.7}} & \fade{602.5} & \fade{-} & \fade{1114.3} & \fade{-} & \fade{-} \\
    & \fade{w/o tuning} & \fade{138.6} & \textbf{\fade{41.2}} & \fade{514.9} & \fade{-} & \fade{1053.3} & \fade{-} & \fade{-} \\
    \cmidrule(r){2-2} \cmidrule(r){3-4} \cmidrule(r){5-6} \cmidrule{7-9}
    \multirow{-5}{*}{\makecell{CrossGET(Ours)}} & \cellcolor{gray!10}w/ tuning & \cellcolor{gray!10}\textbf{143.1}$_{\color{red}\downarrow 2.5}$ & \cellcolor{gray!10}\textbf{41.9}$_{\color{red}\downarrow 0.9}$ & \cellcolor{gray!10}413.9$_{\color{ForestGreen}\downarrow 52\%}$ & \cellcolor{gray!10}104.5$_{\color{ForestGreen}\uparrow 94\%}$ & \cellcolor{gray!10}822.0$_{\color{ForestGreen}\downarrow 40\%}$ & \cellcolor{gray!10}30.5$_{\color{ForestGreen}\uparrow 37\%}$ & \cellcolor{gray!10}84.1$_{\color{ForestGreen}\uparrow 60\%}$ \\
    \bottomrule
  \end{tabular}
  \label{table blip2 small caption}

\end{table*}

Experimental results on BLIP2-OPT2.7B \cite{li2023blip} are listed in Table \ref{table blip2 small caption}, which demonstrates similarly promising performance of \textit{CrossGET} as on BLIP2-OPT6.7B. Note that the performance of the original model we tested locally is slightly lower than the results reported in the original paper.

\subsection{Experiments with BLIP2 about Where to Reduce Tokens}
\label{section apply}

We conduct experiments on directly ensembling tokens on OPT. To elaborate, directly ensembling tokens on OPT leads to fewer tokens stored in the KV cache. Therefore, during each generation step, a smaller number of previous tokens’ KV cache will attend to the current token and thus need less computational cost for Self-Attentions.

\begin{table*}[h]
\captionsetup{font={small}}
\small
\centering
\caption{Accelerate multimodal LLM BLIP2-OPT on the COCO Caption dataset of the Image Caption task.}
\vspace{-6pt}
    \begin{tabular}{l l l l}
    \toprule
    Method & Where to reduce tokens & CIDEr & GFLOPs \\
    \cmidrule(r){1-4}
    BLIP2-OPT2.7B & / & 145.6 & 854.2 \\
    BLIP2-OPT2.7B with CrossGET	& On ViT and Q-Former & 143.1 & 413.9 \\
    BLIP2-OPT2.7B with CrossGET & On ViT and LLM & 142.4$_{\color{red}\downarrow 0.7}$ & 417.7 \\
    \cmidrule(r){1-4}
    BLIP2-OPT6.7B & / & 144.5 & 1042.6 \\
    BLIP2-OPT6.7B with CrossGET & On ViT and Q-Former & 143.1 & 558.2 \\
    BLIP2-OPT6.7B with CrossGET & On ViT and LLM & 143.5$_{\color{ForestGreen}\uparrow 0.4}$ & 566.1 \\
    \bottomrule
    \end{tabular}
\label{table where to apply}
\end{table*}

An intriguing finding from Table \ref{table where to apply} is that the performance of OPT is positively affected by directly ensembling tokens on the relatively larger OPT6.7B model while negatively affected by the same setting on the relatively smaller OPT2.7B model. A possible explanation for the contrasting behaviors is: 

\begin{itemize}
    \item To accelerate OPT, the smaller 2.7B model is more vulnerable to the disturbance brought by the token ensemble within the model (note that the OPT model is frozen, so it cannot adapt its weights of parameters when the number of tokens is getting smaller). Therefore, applying CrossGET on Q-Former is a better setting so that the OPT model is accelerated by taking fewer input tokens. 
    \item The larger 6.7B model is more resilient to the disturbance brought by the token ensemble within the model. Moreover, ensembling tokens within the OPT model can help preserve the tokens’ information as much as possible (if the number of tokens is reduced in the preceding Q-Former, the lost information due to the ensemble operation will be inaccessible to the succeeding OPT model). 
\end{itemize}

Moreover, the experiments also indicate that after ensembling tokens, at least two competing factors determine the extent of the performance affected: 1) performance increases due to preserving more tokens’ information within the OPT model. 2) performance decreases depending on frozen OPT’s resilience to the disturbance brought by token ensemble within the model.

\subsection{Evaluation at Different Reduction Ratios without Training}
\label{section no training}

Exhaustive experimental results at different reduction ratios \textbf{\textit{without}} training are listed in Table \ref{table no train}. 

\begin{table*}[!p]
\setlength{\tabcolsep}{30mm}
\captionsetup{font={small}}
\small
\centering
\setlength{\tabcolsep}{0.8mm}	
\caption{Experimental results at different reduction ratios for BLIP on the NVLR2 dataset of the Visual Reasoning task \textbf{\textit{without}} training.} 
    \begin{tabular}{l @{\hspace{5\tabcolsep}} l l @{\hspace{5\tabcolsep}} l @{\hspace{5\tabcolsep}}l l @{\hspace{5\tabcolsep}} l}
    \toprule
    Approach & \multicolumn{2}{c}{\textcolor[RGB]{172, 172, 234}{Test Acc}} & Drop & \multicolumn{2}{c}{\textcolor[RGB]{241, 205, 122}{GFLOPs}} & Reduction \\
    \cmidrule(r){1-1} \cmidrule(r){2-4} \cmidrule{5-7}
    BLIP \cite{li2022blip} & 83.38 & \begin{tikzpicture}\filldraw[lavender] (0,-0.1) rectangle (2,0.12);\end{tikzpicture} & \fade{-} & 132.54 & \begin{tikzpicture}\filldraw[LemonChiffon] (0,-0.1) rectangle (2,0.12);\end{tikzpicture} & \fade{-} \\
    \cmidrule(r){1-1} \cmidrule(r){2-4} \cmidrule{5-7}
    & 83.34 & \begin{tikzpicture}\filldraw[lavender] (0,-0.1) rectangle (2*0.8334/0.8338,0.12);\end{tikzpicture} & \fade{-0.04} & 136.92 & \begin{tikzpicture}\filldraw[LemonChiffon] (0,-0.1) rectangle (2*136.92/132.54,0.12);\end{tikzpicture} & \fade{0.97x} \\
    & 83.32 & \begin{tikzpicture}\filldraw[lavender] (0,-0.1) rectangle (2*0.8332/0.8338,0.12);\end{tikzpicture} & \fade{-0.06} & 135.18 & \begin{tikzpicture}\filldraw[LemonChiffon] (0,-0.1) rectangle (2*135.18/132.54,0.12);\end{tikzpicture} & \fade{0.98x} \\
    & 83.40 & \begin{tikzpicture}\filldraw[lavender] (0,-0.1) rectangle (2*0.8340/0.8338,0.12);\end{tikzpicture} & \fade{+0.02} & 133.43 & \begin{tikzpicture}\filldraw[LemonChiffon] (0,-0.1) rectangle (2*133.43/132.54,0.12);\end{tikzpicture} & \fade{0.99x}  \\
    & 83.22 & \begin{tikzpicture}\filldraw[lavender] (0,-0.1) rectangle (2*0.8322/0.8338,0.12);\end{tikzpicture} & \fade{-0.16} & 131.69 & \begin{tikzpicture}\filldraw[LemonChiffon] (0,-0.1) rectangle (2*131.69/132.54,0.12);\end{tikzpicture} & \fade{1.01x} \\
    & 83.17 & \begin{tikzpicture}\filldraw[lavender] (0,-0.1) rectangle (2*0.8317/0.8338,0.12);\end{tikzpicture} & \fade{-0.21} & 129.96 & \begin{tikzpicture}\filldraw[LemonChiffon] (0,-0.1) rectangle (2*129.96/132.54,0.12);\end{tikzpicture} & \fade{1.02x} \\
    & 83.02 & \begin{tikzpicture}\filldraw[lavender] (0,-0.1) rectangle (2*0.8302/0.8338,0.12);\end{tikzpicture} & \fade{-0.36} & 128.23 & \begin{tikzpicture}\filldraw[LemonChiffon] (0,-0.1) rectangle (2*128.23/132.54,0.12);\end{tikzpicture} & \fade{1.03x} \\
    & 83.08 & \begin{tikzpicture}\filldraw[lavender] (0,-0.1) rectangle (2*0.8308/0.8338,0.12);\end{tikzpicture} & \fade{-0.30} & 126.50 & \begin{tikzpicture}\filldraw[LemonChiffon] (0,-0.1) rectangle (2*126.50/132.54,0.12);\end{tikzpicture} & \fade{1.05x} \\
    & 82.99 & \begin{tikzpicture}\filldraw[lavender] (0,-0.1) rectangle (2*0.8299/0.8338,0.12);\end{tikzpicture} & \fade{-0.39} & 124.78 & \begin{tikzpicture}\filldraw[LemonChiffon] (0,-0.1) rectangle (2*124.78/132.54,0.12);\end{tikzpicture} & \fade{1.06x} \\
    & 82.88 & \begin{tikzpicture}\filldraw[lavender] (0,-0.1) rectangle (2*0.8288/0.8338,0.12);\end{tikzpicture} & \fade{-0.50} & 123.07 & \begin{tikzpicture}\filldraw[LemonChiffon] (0,-0.1) rectangle (2*123.07/132.54,0.12);\end{tikzpicture} & \fade{1.08x} \\
    & 82.81 & \begin{tikzpicture}\filldraw[lavender] (0,-0.1) rectangle (2*0.8281/0.8338,0.12);\end{tikzpicture} & \fade{-0.57} & 121.36 & \begin{tikzpicture}\filldraw[LemonChiffon] (0,-0.1) rectangle (2*121.36/132.54,0.12);\end{tikzpicture} & \fade{1.09x} \\
    & 82.85 & \begin{tikzpicture}\filldraw[lavender] (0,-0.1) rectangle (2*0.8285/0.8338,0.12);\end{tikzpicture} & \fade{-0.53} & 119.65 & \begin{tikzpicture}\filldraw[LemonChiffon] (0,-0.1) rectangle (2*119.65/132.54,0.12);\end{tikzpicture} & \fade{1.11x} \\
    & 82.66 & \begin{tikzpicture}\filldraw[lavender] (0,-0.1) rectangle (2*0.8266/0.8338,0.12);\end{tikzpicture} & \fade{-0.72} & 117.95 & \begin{tikzpicture}\filldraw[LemonChiffon] (0,-0.1) rectangle (2*117.95/132.54,0.12);\end{tikzpicture} & \fade{1.12x} \\
    & 82.48 & \begin{tikzpicture}\filldraw[lavender] (0,-0.1) rectangle (2*0.8248/0.8338,0.12);\end{tikzpicture} & \fade{-0.90} & 116.25 & \begin{tikzpicture}\filldraw[LemonChiffon] (0,-0.1) rectangle (2*116.25/132.54,0.12);\end{tikzpicture} & \fade{1.14x} \\
    & 82.20 & \begin{tikzpicture}\filldraw[lavender] (0,-0.1) rectangle (2*0.8220/0.8338,0.12);\end{tikzpicture} & \fade{-1.18} & 114.56 & \begin{tikzpicture}\filldraw[LemonChiffon] (0,-0.1) rectangle (2*114.56/132.54,0.12);\end{tikzpicture} & \fade{1.16x} \\
    & 82.30 & \begin{tikzpicture}\filldraw[lavender] (0,-0.1) rectangle (2*0.8230/0.8338,0.12);\end{tikzpicture} & \fade{-1.08} & 112.87 & \begin{tikzpicture}\filldraw[LemonChiffon] (0,-0.1) rectangle (2*112.87/132.54,0.12);\end{tikzpicture} & \fade{1.17x} \\
    & 82.02 & \begin{tikzpicture}\filldraw[lavender] (0,-0.1) rectangle (2*0.8202/0.8338,0.12);\end{tikzpicture} & \fade{-1.36} & 111.19 & \begin{tikzpicture}\filldraw[LemonChiffon] (0,-0.1) rectangle (2*111.19/132.54,0.12);\end{tikzpicture} & \fade{1.19x} \\
    & 81.67 & \begin{tikzpicture}\filldraw[lavender] (0,-0.1) rectangle (2*0.8167/0.8338,0.12);\end{tikzpicture} & \fade{-1.71} & 109.51 & \begin{tikzpicture}\filldraw[LemonChiffon] (0,-0.1) rectangle (2*109.51/132.54,0.12);\end{tikzpicture} & \fade{1.21x} \\
    & 81.90 & \begin{tikzpicture}\filldraw[lavender] (0,-0.1) rectangle (2*0.8190/0.8338,0.12);\end{tikzpicture} & \fade{-1.48} & 107.84 & \begin{tikzpicture}\filldraw[LemonChiffon] (0,-0.1) rectangle (2*107.84/132.54,0.12);\end{tikzpicture} & \fade{1.23x} \\
    & 81.75 & \begin{tikzpicture}\filldraw[lavender] (0,-0.1) rectangle (2*0.8175/0.8338,0.12);\end{tikzpicture} & \fade{-1.63} & 106.17 & \begin{tikzpicture}\filldraw[LemonChiffon] (0,-0.1) rectangle (2*106.17/132.54,0.12);\end{tikzpicture} & \fade{1.25x} \\
    & 81.63 & \begin{tikzpicture}\filldraw[lavender] (0,-0.1) rectangle (2*0.8163/0.8338,0.12);\end{tikzpicture} & \fade{-1.75} & 104.51 & \begin{tikzpicture}\filldraw[LemonChiffon] (0,-0.1) rectangle (2*104.51/132.54,0.12);\end{tikzpicture} & \fade{1.27x} \\
    & 81.47 & \begin{tikzpicture}\filldraw[lavender] (0,-0.1) rectangle (2*0.8147/0.8338,0.12);\end{tikzpicture} & \fade{-1.91} & 102.84 & \begin{tikzpicture}\filldraw[LemonChiffon] (0,-0.1) rectangle (2*102.84/132.54,0.12);\end{tikzpicture} & \fade{1.29x} \\
    & 81.43 & \begin{tikzpicture}\filldraw[lavender] (0,-0.1) rectangle (2*0.8143/0.8338,0.12);\end{tikzpicture} & \fade{-1.95} & 101.19 & \begin{tikzpicture}\filldraw[LemonChiffon] (0,-0.1) rectangle (2*101.19/132.54,0.12);\end{tikzpicture} & \fade{1.31x} \\
    & 81.29 & \begin{tikzpicture}\filldraw[lavender] (0,-0.1) rectangle (2*0.8129/0.8338,0.12);\end{tikzpicture} & \fade{-2.09} & 99.54  & \begin{tikzpicture}\filldraw[LemonChiffon] (0,-0.1) rectangle (2*99.54/132.54,0.12);\end{tikzpicture}  & \fade{1.33x} \\
    & 80.93 & \begin{tikzpicture}\filldraw[lavender] (0,-0.1) rectangle (2*0.8093/0.8338,0.12);\end{tikzpicture} & \fade{-2.45} & 97.89  & \begin{tikzpicture}\filldraw[LemonChiffon] (0,-0.1) rectangle (2*97.89/132.54,0.12);\end{tikzpicture}  & \fade{1.35x} \\
    & 80.87 & \begin{tikzpicture}\filldraw[lavender] (0,-0.1) rectangle (2*0.8087/0.8338,0.12);\end{tikzpicture} & \fade{-2.51} & 96.25  & \begin{tikzpicture}\filldraw[LemonChiffon] (0,-0.1) rectangle (2*96.25/132.54,0.12);\end{tikzpicture}  & \fade{1.38x} \\
    & 80.93 & \begin{tikzpicture}\filldraw[lavender] (0,-0.1) rectangle (2*0.8093/0.8338,0.12);\end{tikzpicture} & \fade{-2.45} & 94.61  & \begin{tikzpicture}\filldraw[LemonChiffon] (0,-0.1) rectangle (2*94.61/132.54,0.12);\end{tikzpicture}  & \fade{1.40x} \\
    & 80.86 & \begin{tikzpicture}\filldraw[lavender] (0,-0.1) rectangle (2*0.8086/0.8338,0.12);\end{tikzpicture} & \fade{-2.52} & 92.98  & \begin{tikzpicture}\filldraw[LemonChiffon] (0,-0.1) rectangle (2*92.98/132.54,0.12);\end{tikzpicture}  & \fade{1.43x} \\
    & 80.68 & \begin{tikzpicture}\filldraw[lavender] (0,-0.1) rectangle (2*0.8068/0.8338,0.12);\end{tikzpicture} & \fade{-2.70} & 91.35  & \begin{tikzpicture}\filldraw[LemonChiffon] (0,-0.1) rectangle (2*91.35/132.54,0.12);\end{tikzpicture}  & \fade{1.45x} \\
    & 80.55 & \begin{tikzpicture}\filldraw[lavender] (0,-0.1) rectangle (2*0.8055/0.8338,0.12);\end{tikzpicture} & \fade{-2.83} & 89.73  & \begin{tikzpicture}\filldraw[LemonChiffon] (0,-0.1) rectangle (2*89.73/132.54,0.12);\end{tikzpicture}  & \fade{1.48x} \\
    & 80.28 & \begin{tikzpicture}\filldraw[lavender] (0,-0.1) rectangle (2*0.8028/0.8338,0.12);\end{tikzpicture} & \fade{-3.10} & 88.12  & \begin{tikzpicture}\filldraw[LemonChiffon] (0,-0.1) rectangle (2*88.12/132.54,0.12);\end{tikzpicture}  & \fade{1.50x} \\
    & 80.35 & \begin{tikzpicture}\filldraw[lavender] (0,-0.1) rectangle (2*0.8035/0.8338,0.12);\end{tikzpicture} & \fade{-3.03} & 86.50  & \begin{tikzpicture}\filldraw[LemonChiffon] (0,-0.1) rectangle (2*86.50/132.54,0.12);\end{tikzpicture}  & \fade{1.53x} \\
    & 80.22 & \begin{tikzpicture}\filldraw[lavender] (0,-0.1) rectangle (2*0.8022/0.8338,0.12);\end{tikzpicture} & \fade{-3.16} & 84.89  & \begin{tikzpicture}\filldraw[LemonChiffon] (0,-0.1) rectangle (2*84.89/132.54,0.12);\end{tikzpicture}  & \fade{1.56x} \\
    & 80.22 & \begin{tikzpicture}\filldraw[lavender] (0,-0.1) rectangle (2*0.8022/0.8338,0.12);\end{tikzpicture} & \fade{-3.16} & 83.29  & \begin{tikzpicture}\filldraw[LemonChiffon] (0,-0.1) rectangle (2*83.29/132.54,0.12);\end{tikzpicture}  & \fade{1.59x} \\
    & 80.38 & \begin{tikzpicture}\filldraw[lavender] (0,-0.1) rectangle (2*0.8038/0.8338,0.12);\end{tikzpicture} & \fade{-3.00} & 81.69  & \begin{tikzpicture}\filldraw[LemonChiffon] (0,-0.1) rectangle (2*81.69/132.54,0.12);\end{tikzpicture}  & \fade{1.62x} \\
    & 80.17 & \begin{tikzpicture}\filldraw[lavender] (0,-0.1) rectangle (2*0.8017/0.8338,0.12);\end{tikzpicture} & \fade{-3.21} & 80.10  & \begin{tikzpicture}\filldraw[LemonChiffon] (0,-0.1) rectangle (2*80.10/132.54,0.12);\end{tikzpicture}  & \fade{1.65x} \\
    & 80.17 & \begin{tikzpicture}\filldraw[lavender] (0,-0.1) rectangle (2*0.8017/0.8338,0.12);\end{tikzpicture} & \fade{-3.21} & 78.51  & \begin{tikzpicture}\filldraw[LemonChiffon] (0,-0.1) rectangle (2*78.51/132.54,0.12);\end{tikzpicture}  & \fade{1.69x} \\
    & 80.30 & \begin{tikzpicture}\filldraw[lavender] (0,-0.1) rectangle (2*0.8030/0.8338,0.12);\end{tikzpicture} & \fade{-3.08} & 76.92  & \begin{tikzpicture}\filldraw[LemonChiffon] (0,-0.1) rectangle (2*76.92/132.54,0.12);\end{tikzpicture}  & \fade{1.72x} \\
    & 80.12 & \begin{tikzpicture}\filldraw[lavender] (0,-0.1) rectangle (2*0.8012/0.8338,0.12);\end{tikzpicture} & \fade{-3.26} & 75.34  & \begin{tikzpicture}\filldraw[LemonChiffon] (0,-0.1) rectangle (2*75.34/132.54,0.12);\end{tikzpicture}  & \fade{1.76x} \\
    & 80.21 & \begin{tikzpicture}\filldraw[lavender] (0,-0.1) rectangle (2*0.8021/0.8338,0.12);\end{tikzpicture} & \fade{-3.17} & 73.76  & \begin{tikzpicture}\filldraw[LemonChiffon] (0,-0.1) rectangle (2*73.76/132.54,0.12);\end{tikzpicture}  & \fade{1.80x} \\
    & 80.02 & \begin{tikzpicture}\filldraw[lavender] (0,-0.1) rectangle (2*0.8002/0.8338,0.12);\end{tikzpicture} & \fade{-3.36} & 72.19  & \begin{tikzpicture}\filldraw[LemonChiffon] (0,-0.1) rectangle (2*72.19/132.54,0.12);\end{tikzpicture}  & \fade{1.84x} \\
    & 79.79 & \begin{tikzpicture}\filldraw[lavender] (0,-0.1) rectangle (2*0.7979/0.8338,0.12);\end{tikzpicture} & \fade{-3.59} & 70.63  & \begin{tikzpicture}\filldraw[LemonChiffon] (0,-0.1) rectangle (2*70.63/132.54,0.12);\end{tikzpicture}  & \fade{1.88x} \\
    & 79.64 & \begin{tikzpicture}\filldraw[lavender] (0,-0.1) rectangle (2*0.7964/0.8338,0.12);\end{tikzpicture} & \fade{-3.74} & 69.07  & \begin{tikzpicture}\filldraw[LemonChiffon] (0,-0.1) rectangle (2*69.07/132.54,0.12);\end{tikzpicture}  & \fade{1.92x} \\
    & 80.02 & \begin{tikzpicture}\filldraw[lavender] (0,-0.1) rectangle (2*0.8002/0.8338,0.12);\end{tikzpicture} & \fade{-3.36} & 67.51  & \begin{tikzpicture}\filldraw[LemonChiffon] (0,-0.1) rectangle (2*67.51/132.54,0.12);\end{tikzpicture}  & \fade{1.96x} \\
    & 79.87 & \begin{tikzpicture}\filldraw[lavender] (0,-0.1) rectangle (2*0.7987/0.8338,0.12);\end{tikzpicture} & \fade{-3.51} & 65.96  & \begin{tikzpicture}\filldraw[LemonChiffon] (0,-0.1) rectangle (2*65.96/132.54,0.12);\end{tikzpicture}  & \fade{2.01x} \\
    & 79.68 & \begin{tikzpicture}\filldraw[lavender] (0,-0.1) rectangle (2*0.7968/0.8338,0.12);\end{tikzpicture} & \fade{-3.70} & 64.60  & \begin{tikzpicture}\filldraw[LemonChiffon] (0,-0.1) rectangle (2*64.60/132.54,0.12);\end{tikzpicture}  & \fade{2.05x} \\
    & 79.32 & \begin{tikzpicture}\filldraw[lavender] (0,-0.1) rectangle (2*0.7932/0.8338,0.12);\end{tikzpicture} & \fade{-4.06} & 63.33  & \begin{tikzpicture}\filldraw[LemonChiffon] (0,-0.1) rectangle (2*63.33/132.54,0.12);\end{tikzpicture}  & \fade{2.09x} \\
    & 79.10 & \begin{tikzpicture}\filldraw[lavender] (0,-0.1) rectangle (2*0.7910/0.8338,0.12);\end{tikzpicture} & \fade{-4.28} & 62.03  & \begin{tikzpicture}\filldraw[LemonChiffon] (0,-0.1) rectangle (2*62.03/132.54,0.12);\end{tikzpicture}  & \fade{2.14x} \\
    & 78.80 & \begin{tikzpicture}\filldraw[lavender] (0,-0.1) rectangle (2*0.7880/0.8338,0.12);\end{tikzpicture} & \fade{-4.58} & 60.78  & \begin{tikzpicture}\filldraw[LemonChiffon] (0,-0.1) rectangle (2*60.78/132.54,0.12);\end{tikzpicture}  & \fade{2.18x} \\
    & 78.85 & \begin{tikzpicture}\filldraw[lavender] (0,-0.1) rectangle (2*0.7885/0.8338,0.12);\end{tikzpicture} & \fade{-4.53} & 59.73  & \begin{tikzpicture}\filldraw[LemonChiffon] (0,-0.1) rectangle (2*59.73/132.54,0.12);\end{tikzpicture}  & \fade{2.22x} \\
    \multirow{-50}{*}{\makecell[l]{CrossGET \\ (\textbf{\textit{without}} training)}} & 78.85 & \begin{tikzpicture}\filldraw[lavender] (0,-0.1) rectangle (2*0.7885/0.8338,0.12);\end{tikzpicture} & \fade{-4.53} & 58.65 & \begin{tikzpicture}\filldraw[LemonChiffon] (0,-0.1) rectangle (2*58.65/132.54,0.12);\end{tikzpicture} & \fade{2.26x} \\
    \bottomrule
  \end{tabular}
  \label{table no train}
\end{table*}

\subsection{Evaluation at Different Reduction Ratios with Training}
\label{section training}

More experimental results at different reduction ratios with training are listed in Table \ref{table train}. 

\begin{table*}[!p]
\setlength{\tabcolsep}{30mm}
\captionsetup{font={small}}
\small
\centering
\setlength{\tabcolsep}{0.8mm}	
\caption{Experimental results at different reduction ratios for BLIP on the NVLR2 dataset of the Visual Reasoning task with training.} 
    \begin{tabular}{l @{\hspace{5\tabcolsep}} l l @{\hspace{5\tabcolsep}} l @{\hspace{5\tabcolsep}}l l @{\hspace{5\tabcolsep}} l}
    \toprule
    Approach & \multicolumn{2}{c}{\textcolor[RGB]{172, 172, 234}{Test Acc}} & Drop & \multicolumn{2}{c}{\textcolor[RGB]{241, 205, 122}{GFLOPs}} & Reduction \\
    \cmidrule(r){1-1} \cmidrule(r){2-4} \cmidrule{5-7}
    BLIP \cite{li2022blip} & 83.38 & \begin{tikzpicture}\filldraw[lavender] (0,-0.1) rectangle (2,0.12);\end{tikzpicture} & \fade{-} & 132.54 & \begin{tikzpicture}\filldraw[LemonChiffon] (0,-0.1) rectangle (2,0.12);\end{tikzpicture} & \fade{-} \\
    \cmidrule(r){1-1} \cmidrule(r){2-4} \cmidrule{5-7}
    & 83.74 & \begin{tikzpicture}\filldraw[lavender] (0,-0.1) rectangle (2*83.74/83.38,0.12);\end{tikzpicture} & \fade{+0.36} & 118.34 & \begin{tikzpicture}\filldraw[LemonChiffon] (0,-0.1) rectangle (2*118.34/132.54,0.12);\end{tikzpicture} & \fade{1.12x} \\
    & 83.31 & \begin{tikzpicture}\filldraw[lavender] (0,-0.1) rectangle (2*83.31/83.38,0.12);\end{tikzpicture} & \fade{-0.07} & 85.27 & \begin{tikzpicture}\filldraw[LemonChiffon] (0,-0.1) rectangle (2*85.27/132.54,0.12);\end{tikzpicture} & \fade{1.55x} \\
    & 83.19 & \begin{tikzpicture}\filldraw[lavender] (0,-0.1) rectangle (2*83.19/83.38,0.12);\end{tikzpicture} & \fade{-0.19}  & 61.09 & \begin{tikzpicture}\filldraw[LemonChiffon] (0,-0.1) rectangle (2*61.09/132.54,0.12);\end{tikzpicture}  &\fade{2.17x} \\
    & 82.28 & \begin{tikzpicture}\filldraw[lavender] (0,-0.1) rectangle (2*82.38/83.38,0.12);\end{tikzpicture} & \fade{-1.10} & 58.95 & \begin{tikzpicture}\filldraw[LemonChiffon] (0,-0.1) rectangle (2*58.95/132.54,0.12);\end{tikzpicture}  & \fade{2.25x} \\
    & 81.33 & \begin{tikzpicture}\filldraw[lavender] (0,-0.1) rectangle (2*81.33/83.38,0.12);\end{tikzpicture} & \fade{-2.05} & 53.25 & \begin{tikzpicture}\filldraw[LemonChiffon] (0,-0.1) rectangle (2*53.25/132.54,0.12);\end{tikzpicture}  & \fade{2.49x} \\
    & 80.67 & \begin{tikzpicture}\filldraw[lavender] (0,-0.1) rectangle (2*80.67/83.38,0.12);\end{tikzpicture} & \fade{-2.71} & 50.14 & \begin{tikzpicture}\filldraw[LemonChiffon] (0,-0.1) rectangle (2*50.14/132.54,0.12);\end{tikzpicture} & \fade{2.64x} \\
    \multirow{-7}{*}{\makecell[l]{CrossGET \\ (with training)}} & 78.19 & \begin{tikzpicture}\filldraw[lavender] (0,-0.1) rectangle (2*78.19/83.38,0.12);\end{tikzpicture} & \fade{-5.19} & 45.11 & \begin{tikzpicture}\filldraw[LemonChiffon] (0,-0.1) rectangle (2*45.11/132.54,0.12,0.12);\end{tikzpicture} & \fade{2.94x} \\
    \bottomrule
  \end{tabular}
  \label{table train}
\end{table*}

\subsection{Re-Evaluation Trained Model at Different Reduction Ratios}
\label{section trained for no train}

Once \textit{CrossGET} has trained a model at a certain compression ratio, a series of models with different performance and computational costs are obtained simultaneously. More specifically, by simply adjusting the number of tokens reduced at inference, it is free to use different models without training based on the desired budget. Table \ref{table trained for no train} provides the relevant experimental results for CLIP model on the Flickr30K dataset of the Image-Text Retrieval task.

\begin{table*}[!p]
\setlength{\tabcolsep}{30mm}
\captionsetup{font={small}}
\small
\centering
\setlength{\tabcolsep}{0.8mm}	
\caption{Experimental results for re-evaluating a model trained by CrossGET (50$\%$ tokens reduced) at different reduction ratios \textbf{\textit{without}} training. } 
    \begin{tabular}{l @{\hspace{5\tabcolsep}} l r@{\hspace{0\tabcolsep}}l @{\hspace{5\tabcolsep}} l @{\hspace{5\tabcolsep}}l l @{\hspace{5\tabcolsep}} l}
    \toprule
    Approach & \multicolumn{3}{c}{\textcolor[RGB]{172, 172, 234}{$\overline{\text{Recall@1}}$ - Trained}} & Change & \multicolumn{2}{c}{\textcolor[RGB]{241, 205, 122}{GFLOPs}} & Increase \\
    \cmidrule(r){1-1} \cmidrule(r){2-5} \cmidrule{6-8}
    CLIP \cite{radford2021learning} & 85.70 & \begin{tikzpicture}\filldraw[MistyRose] (0,-0.1) rectangle (2*0.15/0.49,0.12);\end{tikzpicture} & & \fade{-0.15} & 20.57 & \begin{tikzpicture}\filldraw[LemonChiffon] (0,-0.1) rectangle (2,0.12);\end{tikzpicture} & \fade{1.71x} \\
    \cmidrule(r){1-1} \cmidrule(r){2-5} \cmidrule{6-8}
    & 85.86 & &\begin{tikzpicture}\filldraw[lavender] (0,-0.1) rectangle (2*0.01/0.49,0.12);\end{tikzpicture} &  \fade{+0.01} & 20.70 & \begin{tikzpicture}\filldraw[LemonChiffon] (0,-0.1) rectangle (2*20.70/20.57,0.12);\end{tikzpicture} & \fade{1.72x} \\
    & 85.79 & \begin{tikzpicture}\filldraw[MistyRose] (0,-0.1) rectangle (2*0.06/0.49,0.12);\end{tikzpicture} & & \fade{-0.06} & 20.48 & \begin{tikzpicture}\filldraw[LemonChiffon] (0,-0.1) rectangle (2*20.48/20.57,0.12);\end{tikzpicture} & \fade{1.70x} \\
    & 85.93 & & \begin{tikzpicture}\filldraw[lavender] (0,-0.1) rectangle (2*0.08/0.49,0.12);\end{tikzpicture} & \fade{+0.08} & 19.90 & \begin{tikzpicture}\filldraw[LemonChiffon] (0,-0.1) rectangle (2*19.90/20.57,0.12);\end{tikzpicture} & \fade{1.65x} \\
    & 85.88 & & \begin{tikzpicture}\filldraw[lavender] (0,-0.1) rectangle (2*0.03/0.49,0.12);\end{tikzpicture} & \fade{+0.03} & 19.32 & \begin{tikzpicture}\filldraw[LemonChiffon] (0,-0.1) rectangle (2*19.32/20.57,0.12);\end{tikzpicture} & \fade{1.60x} \\
    & 86.06 & & \begin{tikzpicture}\filldraw[lavender] (0,-0.1) rectangle (2*0.21/0.49,0.12);\end{tikzpicture} & \fade{+0.21} & 18.74 & \begin{tikzpicture}\filldraw[LemonChiffon] (0,-0.1) rectangle (2*18.74/20.57,0.12);\end{tikzpicture} & \fade{1.56x} \\
    & 86.19 & & \begin{tikzpicture}\filldraw[lavender] (0,-0.1) rectangle (2*0.34/0.49,0.12);\end{tikzpicture} & \fade{+0.34} & 18.17 & \begin{tikzpicture}\filldraw[LemonChiffon] (0,-0.1) rectangle (2*18.17/20.57,0.12);\end{tikzpicture} & \fade{1.51x} \\
    & 86.34 & & \begin{tikzpicture}\filldraw[lavender] (0,-0.1) rectangle (2*0.49/0.49,0.12);\end{tikzpicture} & \fade{+0.49} & 17.60 & \begin{tikzpicture}\filldraw[LemonChiffon] (0,-0.1) rectangle (2*17.60/20.57,0.12);\end{tikzpicture} & \fade{1.46x} \\
    & 86.15 & & \begin{tikzpicture}\filldraw[lavender] (0,-0.1) rectangle (2*0.30/0.49,0.12);\end{tikzpicture} & \fade{+0.30} & 17.03 & \begin{tikzpicture}\filldraw[LemonChiffon] (0,-0.1) rectangle (2*17.03/20.57,0.12);\end{tikzpicture} & \fade{1.41x} \\
    & 86.21 & & \begin{tikzpicture}\filldraw[lavender] (0,-0.1) rectangle (2*0.36/0.49,0.12);\end{tikzpicture} & \fade{+0.36} & 16.46 & \begin{tikzpicture}\filldraw[LemonChiffon] (0,-0.1) rectangle (2*16.46/20.57,0.12);\end{tikzpicture} & \fade{1.37x} \\
    & 86.33 & & \begin{tikzpicture}\filldraw[lavender] (0,-0.1) rectangle (2*0.48/0.49,0.12);\end{tikzpicture} & \fade{+0.48} & 15.90 & \begin{tikzpicture}\filldraw[LemonChiffon] (0,-0.1) rectangle (2*15.90/20.57,0.12);\end{tikzpicture} & \fade{1.32x} \\
    & 86.19 & & \begin{tikzpicture}\filldraw[lavender] (0,-0.1) rectangle (2*0.34/0.49,0.12);\end{tikzpicture} & \fade{+0.34} & 15.33 & \begin{tikzpicture}\filldraw[LemonChiffon] (0,-0.1) rectangle (2*15.33/20.57,0.12);\end{tikzpicture} & \fade{1.27x} \\
    & 86.29 & & \begin{tikzpicture}\filldraw[lavender] (0,-0.1) rectangle (2*0.49/0.44,0.12);\end{tikzpicture} & \fade{+0.44} & 14.77 & \begin{tikzpicture}\filldraw[LemonChiffon] (0,-0.1) rectangle (2*14.77/20.57,0.12);\end{tikzpicture} & \fade{1.23x} \\
    & 86.06 & & \begin{tikzpicture}\filldraw[lavender] (0,-0.1) rectangle (2*0.21/0.49,0.12);\end{tikzpicture} & \fade{+0.21} & 14.22 & \begin{tikzpicture}\filldraw[LemonChiffon] (0,-0.1) rectangle (2*14.22/20.57,0.12);\end{tikzpicture} & \fade{1.18x} \\
    & 86.01 & & \begin{tikzpicture}\filldraw[lavender] (0,-0.1) rectangle (2*0.16/0.49,0.12);\end{tikzpicture} & \fade{+0.16} & 13.66 & \begin{tikzpicture}\filldraw[LemonChiffon] (0,-0.1) rectangle (2*13.66/20.57,0.12);\end{tikzpicture} & \fade{1.13x} \\
    & 86.08 & & \begin{tikzpicture}\filldraw[lavender] (0,-0.1) rectangle (2*0.23/0.49,0.12);\end{tikzpicture} & \fade{+0.23} & 13.11 & \begin{tikzpicture}\filldraw[LemonChiffon] (0,-0.1) rectangle (2*13.11/20.57,0.12);\end{tikzpicture} & \fade{1.09x} \\
    \multirow{-16}{*}{\makecell[l]{CrossGET \\ (re-evaluate \\ \textbf{\textit{without}} training)}} & 86.20 & & \begin{tikzpicture}\filldraw[lavender] (0,-0.1) rectangle (2*0.35/0.49,0.12);\end{tikzpicture} & \fade{+0.35} & 12.56 & \begin{tikzpicture}\filldraw[LemonChiffon] (0,-0.1) rectangle (2*12.56/20.57,0.12);\end{tikzpicture} & \fade{1.04x} \\
    \cmidrule(r){1-1} \cmidrule(r){2-5} \cmidrule{6-8}
    CrossGET (with training) & 85.85 & & \begin{tikzpicture}\filldraw[lavender] (0,-0.1) rectangle (2*0.00/0.49,0.12);\end{tikzpicture} & - & 12.04 & \begin{tikzpicture}\filldraw[LemonChiffon] (0,-0.1) rectangle (2*12.04/20.57,0.12);\end{tikzpicture} & - \\
    \bottomrule
  \end{tabular}
  \label{table trained for no train}
\end{table*}

\newpage

\section{Supplementary Related Works}
\label{appendix related works}

\paragraph{Token Reduction} Prior works have advanced token reduction in unimodal scenarios, such as for vision \cite{chen2021chasing, rao2021dynamicvit, su2022vitas, chavan2022vision, liang2022not, yin2022vit, liang2022expediting, bolya2022tome} or language \cite{goyal2020power, kim2020length, kim2022learned, lassance2021study}. \textit{CrossGET} emerges as one of the pioneering efforts in token ensemble frameworks for multimodal scenarios. Additionally, It is one of the few approaches requiring no extra learnable parameters aside from negligible cross tokens. Although ToMe \cite{bolya2022tome} also does not require learnable parameters, it is limited to unimodal scenarios. For the convenience of parallelizability, it adopts a bipartite matching method that delivers relatively unreliable token-matching results.

\textbf{Multimodal Transformer Acceleration} \ A few studies have tried to accelerate multimodal Transformers. \citet{gan2022playing} investigates unstructured pruning, discovering that winning tickets \cite{frankle2018lottery} also exist in multimodal Transformers. By structured pruning, UPop \cite{pmlr-v202-shi23e} proposes that small vision-language models can be unifiedly searched within large ones and then progressively pruned. DistillVLM \cite{fang2021compressing} and EfficientVLM \cite{wang2022efficientvlm} suggest knowledge distillation to mimic the distribution of large vision-language models. MiniVLM \cite{wang2020minivlm} employs lightweight networks for its construction. AWQ \cite{lin2023awq} applies weight-only quantization on multimodal Transformers. \textit{CrossGET} achieves acceleration through ensembling tokens, which is orthogonal to these existing strategies by shrinking model parameters. TRIPS \cite{jiang2022trips} utilizes text information for unidirectional guidance in reducing image patches and is limited to modality-independent models. In contrast, \textit{CrossGET} enables bidirectional learning of guidance information between modalities and applies to both modality-independent and modality-dependent models. Appendix \ref{appendix trips} provides more comparisons and analyses on TRIPS.

\paragraph{Parameter-Efficient Fine-Tuning} Parameter-efficient fine-tuning aims to reduce the number of learnable parameters during fine-tuning. It primarily encompasses adapters \cite{houlsby2019parameter, sung2022vl}, prompt tuning \cite{li2021prefix, khattak2022maple}, low-rank adaptation \cite{hu2021lora, hyeon2021fedpara}, parameter sharing \cite{lan2019albert, shi2021multi}, dropout \cite{fan2019reducing, shi2022heuristic} and their combinations \cite{he2021towards, karimi2021compacter}. LST \cite{sung2022lst} suggests a side tuning for enhanced memory efficiency. In multimodal scenarios, LiteVL \cite{chen2022litevl} proposes to inherit image-text pre-trained weights with some slight modifications to quickly adapt to video-text tasks without heavy pre-training, thereby reducing the training cost. While parameter-efficient fine-tuning enhances efficiency in the fine-tuning phase, it does not accelerate model inference. Conversely, \textit{CrossGET} mainly focuses on improving efficiency during inference, and accordingly, the model inference can be significantly accelerated.

\newpage

\section{Supplementary Methodology Details}

\subsection{Examples for Demonstrating Token Matching}
\label{section example token matching}

\paragraph{Optimal Objective Function and Solution} For example, when the number of tokens in total is $N=4$, and the number of tokens to be reduced is $r=2$, by verifying and comparing all possible token-matching results, the optimal objective function for case1 in Figure \ref{figure match} can be obtained:
\begin{equation}
\label{optimal objective function case1}
    S_1^* = \mathcal{D}(\bm{T}_1, \bm{T}_4) + \mathcal{D}(\bm{T}_2, \bm{T}_3) = 0.5 + 0.6 = 1.1,
\end{equation}
and the corresponding optimal solution for token matching can be determined as
\begin{equation}
\label{optimal solution case1}
    \bm{P}_1^*  = \{(\bm{T}_1, \bm{T}_4), (\bm{T}_2, \bm{T}_3) \}.
\end{equation}
Similarly, the optimal objective function for case2 (inverted case1) in Figure \ref{figure match} is
\begin{equation}
\label{optimal objective function case2}
    S_2^* = \mathcal{D}(\bm{T}_1, \bm{T}_3) + \mathcal{D}(\bm{T}_2, \bm{T}_4) = 0.9 + 0.8 = 1.7
\end{equation}
and the corresponding optimal solution for token matching is
\begin{equation}
\label{optimal solution case2}
    \bm{P}_2^*  = \{(\bm{T}_1, \bm{T}_3), (\bm{T}_2, \bm{T}_4) \}.
\end{equation}

\paragraph{Revisiting Bipartite Soft Matching} ToMe \cite{bolya2022tome} suggests a non-iterative \textit{bipartite soft matching} ensure parallelizability, which divides tokens into two disjoint sets alternately, for each token in the first set calculates the maximum similarity from it to each token in the other set, and the token pairs with the highest similarities will be merged. 

Take case1 in Figure \ref{figure match} as an example, tokens are firstly divided into $\{\bm{T}_1, \bm{T}_3\}$ and $\{\bm{T}_2, \bm{T}_4\}$, then the maximum similarity from $\bm{T}_1$ to $\{\bm{T}_2, \bm{T}_4\}$ is $\mathcal{D}(\bm{T}_1, \bm{T}_4)=0.5$ and from $\bm{T}_3$ to $\{\bm{T}_2, \bm{T}_4\}$ is $\mathcal{D}(\bm{T}_3, \bm{T}_2)=0.6$. Therefore, the optimal objective function in Eq.\ref{optimal objective function case1} and optimal solution in Eq.\ref{optimal solution case1} are achieved:
\begin{equation}
    S_1^B = \mathcal{D}(\bm{T}_1, \bm{T}_4) + \mathcal{D}(\bm{T}_2, \bm{T}_3) = S_1^*, \quad \bm{P}_1^B  = \{(\bm{T}_1, \bm{T}_4), (\bm{T}_2, \bm{T}_3) \} = \bm{P}_1^*
\end{equation}
However, for case2 (inverted case1), \textit{bipartite soft matching} leads to a worse solution: tokens are firstly divided into $\{\bm{T}_1, \bm{T}_3\}$ and $\{\bm{T}_2, \bm{T}_4\}$, then the maximum similarity from $\bm{T}_1$ to $\{\bm{T}_2, \bm{T}_4\}$ is $\mathcal{D}(\bm{T}_1, \bm{T}_2)=0.6$ and from $\bm{T}_3$ to $\{\bm{T}_2, \bm{T}_4\}$ is $\mathcal{D}(\bm{T}_3, \bm{T}_4)=0.7$. Therefore, the optimal objective function in Eq.\ref{optimal objective function case2} and optimal solution in Eq.\ref{optimal solution case2} are not achieved:
\begin{equation}
    S_2^{B} = \mathcal{D}(\bm{T}_1, \bm{T}_2) + \mathcal{D}(\bm{T}_3, \bm{T}_4) < S_2^*, \quad \bm{P}_2^{B}  = \{(\bm{T}_1, \bm{T}_2), (\bm{T}_3, \bm{T}_4) \} \neq \bm{P}_2^*
\end{equation}
This is attributed to the design of \textit{bipartite soft matching} that for the convenience of ensuring parallelizability, each token only takes into account the similarity with half but not all other tokens, and the method degrades when tokens with high similarity are not divided into different sets.

\paragraph{Shifting to Complete-Graph Soft Matching} An approximate algorithm \textit{complete-graph soft matching} is proposed to tackle the above challenge. It enables each token to take into account the similarity with all other tokens while avoiding introducing iterative and non-parallelizable operations. 

Take case2 in Figure \ref{figure match} as an example, all tokens $\bm{T} = \{\bm{T}_1, \bm{T}_2, \bm{T}_3, \bm{T}_4\} $ are sorted in descending order according to their maximum similarity to other tokens: $\bm{T}' = \{\bm{T}_1, \bm{T}_3, \bm{T}_2, \bm{T}_4\}$. After adding the dependency mask, the maximum similarity from top priority source token candidate $\bm{T}_1$ to its destination tokens $\{\bm{T}_3, \bm{T}_2, \bm{T}_4\}$ is $\mathcal{D}(\bm{T}_1, \bm{T}_3)=0.9$, from second priority source token candidate $\bm{T}_3$ to its destination tokens $\{\bm{T}_2, \bm{T}_4\}$ is $\mathcal{D}(\bm{T}_3, \bm{T}_4)=0.7$, and from third priority source token candidate $\bm{T}_2$ to its destination token $\bm{T}_4$ is $\mathcal{D}(\bm{T}_2, \bm{T}_4)=0.8$. The source token candidates among them corresponding to the two largest similarities are selected as the source token set $\bm{T}_s = \{\bm{T}_1, \bm{T}_2\} $ while remaining tokens form the destination token set $\bm{T}_D = \{\bm{T}_3, \bm{T}_4\}$. Then the maximum similarity from $\bm{T}_1$ to $\{\bm{T}_3, \bm{T}_4\}$ is $\mathcal{D}(\bm{T}_1, \bm{T}_3)=0.9$ and from $\bm{T}_2$ to $\{\bm{T}_3, \bm{T}_4\}$ is $\mathcal{D}(\bm{T}_2, \bm{T}_4)=0.8$. Therefore, the optimal objective function in Eq.\ref{optimal objective function case2} and optimal solution in Eq.\ref{optimal solution case2} are achieved:
\begin{equation}
    S_2^{C} = \mathcal{D}(\bm{T}_1, \bm{T}_3) + \mathcal{D}(\bm{T}_2, \bm{T}_4) = S_2^*, \quad \bm{P}_2^{C}  = \{(\bm{T}_1, \bm{T}_3), (\bm{T}_2, \bm{T}_4) \} = \bm{P}_2^*.
\end{equation}
Similarly, it can also be verified that \textit{complete-graph soft matching} achieves the optimal objective function in Eq.\ref{optimal objective function case1} and optimal solution in Eq.\ref{optimal solution case1} for case1 in Figure \ref{figure match}.

\newpage
\subsection{Algorithm Implementation of Complete-Graph Soft Matching}
\label{section algorithm cgsm}

\begin{algorithm}[H] 
\caption{Complete-Graph Soft Matching}
\label{algorithm cgsm}
\SetAlgoLined
\KwIn{Number of tokens $N$, number of tokens to be reduced $r$, original tokens $\bm{T} = \{\bm{T}_i\}_{i=1}^{N}$ and their corresponding keys $\bm{K} = \{\bm{K}_i\}_{i=1}^{N}$ where $|\bm{T}|=|\bm{K}|=N$} 

\KwOut{Reduced tokens $\bm{T}^{\star} = \{\bm{T}_i^{\star}\}_{i=1}^{N-r}$ where $|\bm{T}^{\star}|=N-r$}

\textcolor{gray}{\texttt{\# Step1:\,\,Calculate the cosine distance $\bm{D}_{ij}$ between the keys of tokens}}

$\bm{D} = \frac{\bm{K}\bm{K}^{\top}}{\Vert \bm{K} \Vert_2^2} + \operatorname{diag}(\underbrace{-\infty, -\infty, \cdot\cdot\cdot, -\infty}_{N})$, \quad $\bm{D} \in \mathbb{R}^{N \times N}$, \, $\operatorname{diag}: \mathbb{R}^{N} \rightarrow \mathbb{R}^{N\times N}$

\textcolor{gray}{\texttt{\# Step2:\,\,Descendingly sort similarity matrix $\bm{D}$ by maximum similarity}}

$\bm{A}^{S} = \operatorname{argsort}(\max \limits_{1 \leq j \leq N}\bm{D}_{ij}) \in \mathbb{R}^{N}$, \quad $\bm{A}^{D} = \operatorname{argsort}(\max \limits_{1 \leq i \leq N}\bm{D}_{ij}) \in \mathbb{R}^{N}$

$\bm{D}^{\star}=\operatorname{sort_{d}}(\operatorname{sort_{s}}(\bm{D}, \bm{A}^{S}), \bm{A}^{D})$, \,  $\operatorname{sort_{s}}: \bm{D}^{\star}_{ij} \leftarrow \bm{D}_{\bm{A}^{S}_ij}$, \, $\operatorname{sort_{d}}: \bm{D}^{\star}_{ij} \leftarrow \bm{D}_{i\bm{A}^{D}_j}$

\textcolor{gray}{\texttt{\# Step3:\,\,Add a lower triangle dependency mask $\bm{M}$}}

$\bm{D}^{\star} = \bm{D}^{\star} + \bm{M}$, \quad $\bm{M}_{ij}= \begin{cases}-\infty & \text { for } i \geq j \\ 0 & \text { for } i < j\end{cases}$

\textcolor{gray}{\texttt{\# Step4:\,\,Pick source tokens $\bm{T}^S$ and destination tokens $\bm{T}^D$ by similarity}}

$\bm{A} = \operatorname{argsort}(\max \limits_{1 \leq j \leq N}\bm{D}^{\star}_{ij}) \in \mathbb{R}^{N} $, \quad $\bm{A}^S = (\bm{A}_i)_{1\leq i \leq r} \in \mathbb{R}^{r}$, \quad $\bm{T}^S = \{\bm{T}_{i}| i \in \bm{A}^S\}$

$\bm{A} = \underset{j \in (\{k\}_{k=1}^{N}\backslash\bm{A}^S)}{\operatorname{argmax}} \bm{D}^{\star}_{ij}  \in \mathbb{R}^{N} $, \quad $\bm{A}^D = (\bm{A}_i)_{i \in \bm{A}^S} \in \mathbb{R}^{r}$, \quad $\bm{T}^D = \{\bm{T}_{i}| i \in \bm{A}^D\}$

\textcolor{gray}{\texttt{\# Step5:\,\,Average source and corresponding destination tokens}}

\textbf{return} $\bm{T}^{\star} = [\bm{T} \backslash (\bm{T}^S \cup \bm{T}^D)] \cup \{\frac{1}{2}(\bm{T}^S_i + \bm{T}^D_i)\}_{i=1}^{r}$ 
\end{algorithm}

Algorithm \ref{algorithm cgsm} is the detailed implementation of the proposed \textit{complete-graph soft matching}. The Step$1\sim5$ in the comments correspond to the Step$1\sim5$ described in Section \ref{Complete-graph soft matching} of the main text. Regarding parallelizability, there are no sequential loops in the algorithm procedure. Therefore, data can be processed in parallel within each step by parallelizable operations (such as \textit{bmm}, \textit{matmul}, \textit{scatter} and \textit{gather} in Pytorch \cite{paszke2019pytorch}).

\subsection{Algorithm Implementation of Cross-Guided Matching and Ensemble}
\label{section algorithm cgem}

\begin{algorithm}[H] 
\caption{Cross-Guided Matching and Ensemble (\textcolor{cyan}{improvements} upon Algorithm \ref{algorithm cgsm})}
\label{algorithm cgem}
\setcounter{AlgoLine}{0}
\SetAlgoLined
\KwIn{Same inputs as Algorithm \ref{algorithm cgsm}, \textcolor{cyan}{plus query of the cross token $\bm{Q}$}} 

\KwOut{Same as Algorithm \ref{algorithm cgsm}}

\textcolor{gray}{\texttt{\# Step1$\sim$3:\,\,Same as Algorithm \ref{algorithm cgsm}}}

\textcolor{gray}{\texttt{\# \textcolor{cyan}{Step4}:\,\,Pick tokens $\bm{T}^S$ and $\bm{T}^D$ by similarity \textcolor{cyan}{and importance}}}

$\,\textcolor{cyan}{\bm{I} = \frac{\bm{K}\bm{Q}^{\top}}{\Vert \bm{K} \Vert_2\Vert \bm{Q} \Vert_2} \in \mathbb{R}^{N}}$ 

$\bm{A} = \operatorname{argsort}(\max \limits_{1 \leq j \leq N}\bm{D}^{\star}_{ij}\, \textcolor{cyan}{- \,\bm{I}}) \in \mathbb{R}^{N} $, \quad $\bm{A}^S = (\bm{A}_i)_{1\leq i \leq r} \in \mathbb{R}^{r}$, \quad $\bm{T}^S = \{\bm{T}_{i}| i \in \bm{A}^S\}$

$\bm{A} = \underset{j \in (\{k\}_{k=1}^{N}\backslash\bm{A}^S)}{\operatorname{argmax}} \bm{D}^{\star}_{ij}  \in \mathbb{R}^{N} $, \quad $\bm{A}^D = (\bm{A}_i)_{i \in \bm{A}^S} \in \mathbb{R}^{r}$, \quad $\bm{T}^D = \{\bm{T}_{i}| i \in \bm{A}^D\}$

\textcolor{gray}{\texttt{\# \textcolor{cyan}{Step5}:\,\,\textcolor{cyan}{Sum weighted} source and corresponding destination tokens}}

$\textcolor{cyan}{\bm{P} = \{(\bm{T}_i^S, \bm{T}_i^D)\}_{i=1}^{r}}$, \quad $\textcolor{cyan}{\bm{W} = \{\operatorname{softmax}(\bm{I}_i, \bm{I}_j)|(\bm{T}_i, \bm{T}_j)\in \bm{P}\}}$

\textbf{return} $\bm{T}^{\star} = [\bm{T} \backslash (\bm{T}^S \cup \bm{T}^D)] \cup \{\textcolor{cyan}{\sum\nolimits_{j=1}^{|\bm{W}_i|}\bm{W}_{ij}\bm{P}_{ij}}\}_{i=1}^{r}$
\end{algorithm}

Algorithm \ref{algorithm cgem} demonstrates how to improve \textit{complete-graph soft matching} by adding \textit{cross-guided matching and ensemble} upon it. It is worth noting that line$7\sim8$ in Algorithm \ref{algorithm cgem} does not imply that only two tokens are in each stack of tokens to be ensembled. This is because different source tokens in $\bm{T}^S$ may have the same destination token in $\bm{T}^D$, which implies that the size of the stack is allowed to be larger than two (in this case, the procedure of ensembling stacks with the different number of tokens can still be implemented by parallelizable operations such as \textit{scatter\_add} in Pytorch).  

\subsection{Sub-optimal Cases for Complete-Graph Soft Matching}
\label{section sub optimal}

Section \ref{Complete-graph soft matching} has already shown the cases that \textit{complete-graph soft matching} achieves optimal matching, and here we provide more analyses on the sub-optimal cases of \textit{complete-graph soft matching}. 

The main sub-optimal cases come from the trade-off between parallelizability and matching accuracy. To achieve parallelizability, the set of source token $\bm{T}^S$ and destination tokens $\bm{T}^D$ have to be disjoint:
\begin{equation}
\label{phi}
\bm{T}^S \cap \bm{T}^D = \phi.
\end{equation}
Otherwise, consider
\begin{equation}
\bm{T}^S \cap \bm{T}^D =\{\bm{T}_x\} \neq \phi, \quad 1 \leq x \leq N
\end{equation}
where $N$ is the number of the original tokens, then
\begin{equation}
    (\exists \bm{T}_i \in \bm{T}^S \,\, \textit{s.t.} \,\, (\bm{T}_i, \bm{T}_x) \in \bm{P}) \land (\exists \bm{T}_j \in \bm{T}^D \,\, \textit{s.t.} \,\, (\bm{T}_x, \bm{T}_j) \in \bm{P}) 
\end{equation}
where $\bm{P} = \{(\bm{T}_i^S, \bm{T}_i^D)\}_{i=1}^{r}$ is the set of the paired tokens to be ensembled, is true. However, there is a computational dependency between merging $\bm{T}_i$ into $\bm{T}_x$ and merging $\bm{T}_x$ into $\bm{T}_j$. The two operations of the merging require iterations and therefore cannot be parallelized.

$\bm{T}^S$ and $\bm{T}^D$ are disjoint (\ie, Eq.\ref{phi} holds) is equivalent to the constraint 
\begin{equation}
\label{equal}
\forall \bm{T}_i \in \bm{T}^S, \bm{T}_i \notin \bm{T}^D
\end{equation}
is satisfied. In the Step$1$ of the Algorithm \ref{algorithm cgsm}, computation is conducted on a complete graph. Therefore $\bm{T}^S$ and $\bm{T}^D$ are joint, and constraint \ref{equal} does not been satisfied. In Step$3$, the added lower triangle dependency mask ensures that source tokens with higher priority (\ie, whose keys have higher maximum cosine similarity to keys of other tokens) will not become targets for other source tokens with lower priority, \ie, a relaxed constraint
\begin{equation}
\forall \bm{T}_i \in \bm{T}^S, \bm{T}_i \notin (\bm{T}^D_j)_{i \leq j \leq N}
\end{equation}
is satisfied. However, the unsatisfied part of the constraint \ref{equal}
\begin{equation}
\label{constraint}
\forall \bm{T}_i \in \bm{T}^S, \bm{T}_i \notin (\bm{T}^D_j)_{1\leq j < i}
\end{equation}
indicates that source tokens with low priority may still become targets for other source tokens with high priority. To further satisfy constraint \ref{constraint}, the line$10$ of Step$4$ in Algorithm \ref{algorithm cgsm} explicitly removes all elements of the source token set from the set of all tokens to construct the set of the destination tokens. And the sub-optimal cases for \textit{complete-graph soft matching} arise when
\begin{equation}
\underset{j \in (\{k\}_{k=1}^{N}\backslash\bm{A}^S)}{\operatorname{argmax}} \bm{D}^{\star}_{ij} \neq \underset{j \in \{k\}_{k=1}^{N}}{\operatorname{argmax}} \bm{D}^{\star}_{ij},
\end{equation} 
which indicates a source token may exist whose closest destination token in $\bm{T}^D$ happens to be another source token in $\bm{T}^S$. For parallelizability, this destination token is removed from $\bm{T}^D$, resulting in the source token can only match the second closest destination token in the set of reduced $\bm{T}^D$.

\subsection{Expectation of Optimal Matching Probability and Complexity Analysis}
\label{section expectation}

\paragraph{Expectation of Optimal Matching Probability} For a token $\bm{T}_i \in \bm{T}$, assume that any other token $\bm{T}_j \in \bm{T}\backslash\{\bm{T}_i\}$ has the same probability of being its optimal destination token, \ie
\begin{equation}
\forall 1 \leq j \leq N , \quad p((\bm{K}_i, \bm{K}_j) = \underset{\substack{1 \leq k \leq N \\ k \neq i }}{\operatorname{argmax}} \,\, s(\bm{K}_i, \bm{K}_k)) = \frac{1}{N - 1}
\end{equation} 
where $s(x,y)$ is a function that calculates cosine similarity between $x$ and $y$.

For \textit{complete-graph soft matching}, in layer $l$ ($1 \leq l \leq L$), supoose $\bm{X} \sim p(x) $ is a discrete random variable about whether a token from $\bm{T}^S$ ($|\bm{T}^S|=r$) can find its optimal destination token in $\bm{T}^D$ ($|\bm{T}^D|=N-lr$), and the probability distribution $p(x)$ is:
\begin{align}
 p(\bm{X} = \text{can}) & = \sum\limits_{1}^{|\bm{T}^D|} p((\bm{K}_i, \bm{K}_j) = \underset{\substack{1 \leq k \leq N+(1-l)r \\ k \neq i }}{\operatorname{argmax}} s(\bm{K}_i, \bm{K}_k)) \\
 & = \frac{N-lr}{N+(1-l)r-1}. \\
 p(\bm{X} = \text{not}) & = 1 - p(\bm{X} = \text{can}).
\end{align} 
Denote $\bm{L} \sim p(l)$ as a discrete random variable ($\bm{L} \perp\!\!\!\perp \bm{X}$) about the current layer number, and
\begin{equation}
\forall 1 \leq l \leq L, \quad p(\bm{L} = l) = \frac{1}{L}
\end{equation} 
Denote $h(\bm{X}, \bm{L})$ as a indicator function
\begin{equation}
h(\bm{X}, \bm{L}) = \begin{cases}1 & \text { for } \bm{X} = \text{can} \\ 0 & \text { for } \bm{X} = \text{not} \end{cases}
\end{equation} 
Then the expectation of a token from $\bm{T}^S$ can find its optimal destination token in $\bm{T}^D$ is
\begin{align}
\mathbb{E}^C & = \mathbb{E}_{XL}\left[h(\bm{X}, \bm{L})\right] = \sum\limits_{l \in \bm{L}}\sum\limits_{x \in \bm{X}}h(x, l)p_{XL}(x,l) \\
& = \sum\limits_{l \in \bm{L}}\sum\limits_{x \in \bm{X}}h(x, l)p_{\bm{X}}(x)p_{\bm{L}}(l) \\
& = \sum\limits_{l = 1}^{L}[1 \cdot p(\bm{X} = \text{can}) + 0 \cdot p(\bm{X} = \text{not})] \frac{1}{L} \\
& = \frac{1}{L} \sum\limits_{l = 1}^{L}\frac{N-lr}{N+(1-l)r-1}
\end{align} 

Similarly, given by \textit{bipartite soft matching} used in ToMe \cite{bolya2022tome}, the expectation of a token from $\bm{T}^S$ ($|\bm{T}^S|=\lceil\frac{N+(1-l)r}{2}\rceil$) can find its optimal destination token in $\bm{T}^D$ ($|\bm{T}^D|=\lfloor\frac{N+(1-l)r}{2}\rfloor$) is 
\begin{equation}
    \mathbb{E}^B = \frac{1}{L} \sum\limits_{l = 1}^{L}\frac{1}{N+(1-l)r-1}\lfloor\frac{N+(1-l)r}{2}\rfloor
\end{equation}

Compare $\mathbb{E}^C$ given by \textit{complete-graph soft matching} with $\mathbb{E}^B$ give by \textit{bipartite soft matching}:
\begin{align}
\mathbb{E}^C - \mathbb{E}^B & = \frac{1}{L} \sum\limits_{l = 1}^{L}\frac{1}{N+(1-l)r-1}(N-lr - \lfloor\frac{N+(1-l)r}{2}\rfloor)\\
& \geq \frac{1}{L} \sum\limits_{l = 1}^{L}\frac{1}{N+(1-l)r-1}(N-lr - \frac{N+(1-l)r}{2}) \\ 
\label{equation compare}
& = \frac{1}{L} [\underbrace{\sum\limits_{l = 1}^{L-1}\frac{1}{N+(1-l)r-1}\frac{N-lr-r}{2}}_{Part1:\,\, 1 \leq l \leq L-1} + \underbrace{\frac{1}{N+(1-L)r-1}\frac{N-Lr-r}{2}}_{Part2:\,\, l=L}]
\end{align}
For part1 in Eq.\ref{equation compare}, since the number of remaining tokens is always a positive integer, we have
\begin{equation}
    N - Lr \geq 1,
\end{equation}
and therefore for $1 \leq l \leq L-1$:
\begin{equation}
    N - lr \geq r + 1 \Leftrightarrow N - lr - r \geq 1 > 0
\end{equation}
always holds. Morever
\begin{equation}
    N + (1 - l)r - 1 = (N - lr - 1) + r > 0
\end{equation}
always holds. Furthermore, we have part1 $ > 0$ always holds, which indicates the expectation given by \textit{complete-graph soft matching} is higher than \textit{bipartite soft matching} except for the last layer.

For part2 in Eq.\ref{equation compare}, \textit{bipartite soft matching} evenly divides tokens into two disjoint sets, and the size of each set is not less than $r$. However, the remaining tokens before the last layer may be less than $2r$. In such a situation, \textit{bipartite soft matching} have to reduce the $r$ to the $r^{\star}$ such that 
\begin{equation}
    N-Lr^{\star} = r^{\star}
\end{equation}
\textit{complete-graph soft matching} follows the same design, and therefore part2 $= 0$ holds. Overall, we have
\begin{equation}
    \mathbb{E}^C - \mathbb{E}^B > 0
\end{equation}
always holds. For example, for a CLIP \cite{radford2021learning} model with 
\begin{equation}
    N = 197,\,\,L=12,\,\,r=16
\end{equation}
used in our experiments, given by \textit{complete-graph soft matching}, the expectation of optimal matching probability for a token from $\bm{T}^S$ is
\begin{equation}
    \mathbb{E}^C = \frac{1}{12} \sum\limits_{l = 1}^{12}\frac{197-l\times 16}{197+(1-l)\times 16-1} \approx 0.78,
\end{equation}
while given by \textit{bipartite soft matching}, the corresponding expectation is 
\begin{equation}
    \mathbb{E}^B = \frac{1}{12} \sum\limits_{l = 1}^{12}\frac{1}{197+(1-l)\times 16-1}\lfloor\frac{197+(1-l)\times16}{2}\rfloor = 0.50 < \mathbb{E}^C
\end{equation}

\subsection{Complexity Analysis for Complete-Graph Soft Matching} 

Since the \textit{sort} and \textit{argsort} operations in Algorithm \ref{algorithm cgsm} and \ref{algorithm cgem} can be solved by algorithms with $\mathcal{O}(N\log{}N)$ complexity such as \textit{QuickSort} \cite{hoare1962quicksort}, the major complexity $\mathcal{O}(N^2)$ comes from the computation of cosine similarities between each pair of tokens. 

A comparison of different matching methods is listed in Table \ref{table complexity}, which demonstrates that as a parallelizable method, \textit{CrossGET} can achieve relatively high expectation of optimal matching probability for a certain token from $\bm{T}^S$ with relatively low complexity.

\begin{table*}[!h]
  \setlength{\tabcolsep}{5 pt}
  \captionsetup{font={small}}
  \small
  \centering
  \caption{\textbf{A comparison of different matching methods.} Denote $N$ as the number of the original tokens, $r$ as the number of tokens to be reduced, and $T$ as the number of iterations for k-means.}
  \begin{tabular}{lccll}
    \toprule
    Method & Iterative & Parallelizable & Expectation of Optimal Matching Probability & Complexity \\
    \midrule
    Greedy Search & Yes & \textcolor{brickred}{\xmark} & $\{1\}$ & $\mathcal{O}(rN^2)$ \\
    K-Means & Yes & \textcolor{brickred}{\xmark} & $[1-\epsilon, 1], \ \lim_{T \rightarrow \infty}\epsilon = 0 $ & $\mathcal{O}(rNT)$ \\
    Random & No & \textcolor{ForestGreen}{\cmark} & $\frac{1}{N-1} \in [0, 0+\epsilon],\ \lim_{N \rightarrow \infty}\epsilon = 0 $ & $\mathcal{O}(r)$ \\
    ToMe \cite{bolya2022tome} & No & \textcolor{ForestGreen}{\cmark} & $\frac{1}{L} \sum\limits_{l = 1}^{L}\frac{1}{N+(1-l)r-1}\lfloor\frac{N+(1-l)r}{2}\rfloor \in [\frac{1}{2}, \frac{1}{2} + \epsilon], \ \lim_{N \rightarrow \infty}\epsilon = 0$ & $\mathcal{O}(N^2)$ \\
    \rowcolor{gray!10}CrossGET (Ours) & No & \textcolor{ForestGreen}{\cmark} & $\frac{1}{L} \sum\limits_{l = 1}^{L}\frac{N-lr}{N+(1-l)r-1} \in [1-\epsilon, 1], \ \lim_{N \rightarrow \infty}\epsilon = 0$ & $\mathcal{O}(N^2)$ \\
    \bottomrule
  \label{table complexity}
  \end{tabular}
\end{table*}


\end{document}